\documentclass[10pt,twocolumn,letterpaper]{article}

\usepackage[pagenumbers]{wacv} %

\usepackage{ifthen}
\newcommand{\kerabase}[1]{\textcolor[RGB]{0,125,255}{\textbf{Kera: }#1}}

\newcommand{\kera}[1]{\ifthenelse{\equal{\showComment}{1}}{\kerabase{#1}}}

\usepackage{amsmath,amssymb,amsfonts,mathcomp}
\usepackage{multirow}
\usepackage{physics}
\usepackage{bm}

\def\cA{{\mathcal{A}}}

\def\cL{{\mathcal{L}}}

\def\cX{{\mathcal{X}}}

\def\bR{{\mathbb{R}}}

\newtheorem{problem}{Problem}
\usepackage[accsupp]{axessibility}  %

\definecolor{wacvblue}{rgb}{0.21,0.49,0.74}
\usepackage[pagebackref,breaklinks,colorlinks,allcolors=wacvblue]{hyperref}

\def\wacvPaperID{516} %
\def\confName{WACV}
\def\confYear{2026}

\title{Matching Semantically Similar Non-Identical Objects}

\author {
    Yusuke Marumo$^{1\dag}$
    \;
    Kazuhiko Kawamoto$^{2\dag}$
    \;
    Satomi Tanaka$^{3\ddag}$
    \;
    Shigenobu Hirano$^{4\ddag}$
    \;
    Hiroshi Kera$^{5\dag}$\thanks{Corresponding Author.}
    \and
    $^{\dag}$Chiba University
    \quad
    $^{\ddag}$Ricoh Company
    \and
    {\tt\small $^1$yusuke.marumo@chiba-u.jp}, 
    {\tt\small $^2$kawa@faculty.chiba-u.jp}, 
    {\tt\small $^3$satomi.st.tanaka@jp.ricoh.com}
    \and
    {\tt\small $^4$shigenobu.hirano@jp.ricoh.com},
    {\tt\small $^5$kera@chiba-u.jp}
}

\begin{document}
\maketitle
\begin{abstract}
    Not identical but similar objects are ubiquitous in our world, ranging from four-legged animals such as dogs and cats to cars of different models and flowers of various colors.
    This study addresses a novel task of matching such non-identical objects at the pixel level. We propose a weighting scheme of descriptors, Semantic Enhancement Weighting (SEW), that incorporates semantic information from object detectors into existing sparse feature matching methods, extending their targets from identical objects captured from different perspectives to semantically similar objects. The experiments show successful matching between non-identical objects in various cases, including in-class design variations, class discrepancy, and domain shifts (e.g., photo vs. drawing and image corruptions). The code is available at \href{https://github.com/Circ-Leaf/NIOM}{https://github.com/Circ-Leaf/NIOM}.
\end{abstract}
\section{Introduction}\label{sec:intro}
We humans are able to find fine-grained and robust visual correspondences between two objects even when they are not necessarily identical or when they are captured in different situations.
Thanks to this ability, for example, we can find out our missing cats in a gloomy backstreet by remembering them basking in the sun or assembling components to build a shelf based on their perspective drawing. 
In computer vision terms, this needs fine-grained matching of feature points between two objects, even with class discrepancy and domain shift (e.g., image corruptions). 

The abovementioned matching has been partially addressed in the literature.
In the feature matching~\cite{SuperGlue, LoFTR, LightGlue, DarkFeat, LiftFeat}, objects to be matched are assumed identical and captured from different perspectives. 
This assumption also enables supervised training using dense correspondence obtained from 3D reconstruction from image pairs.
Recently, Transformer-based dense matching methods have shown great performance~\cite{LoFTR, sparse_ncnet, patch2pix, cotr, roma, SGAM, ASpanFormer, ASTR, EDM}, and there are also light-weight sparse matching methods, which run in real time with a slight cost in accuracy~\cite{SuperGlue, imp, multiviewmatching, GlueStick, DarkFeat,sceneaware, OmniGlue, mambaglue}. 
In contrast, semantic correspondence models~\cite{semantic_correspondence_survey, GeoAwareSC, diffhyper, sd4match, CNNGeometric} focus on matching semantically similar parts between objects~(e.g., tires of two bikes). They are trained on datasets with sparse part-to-part correspondence annotation and thus do not offer fine-grained matching.
\begin{figure}[tb]
  \centering
  \includegraphics[width=\linewidth]{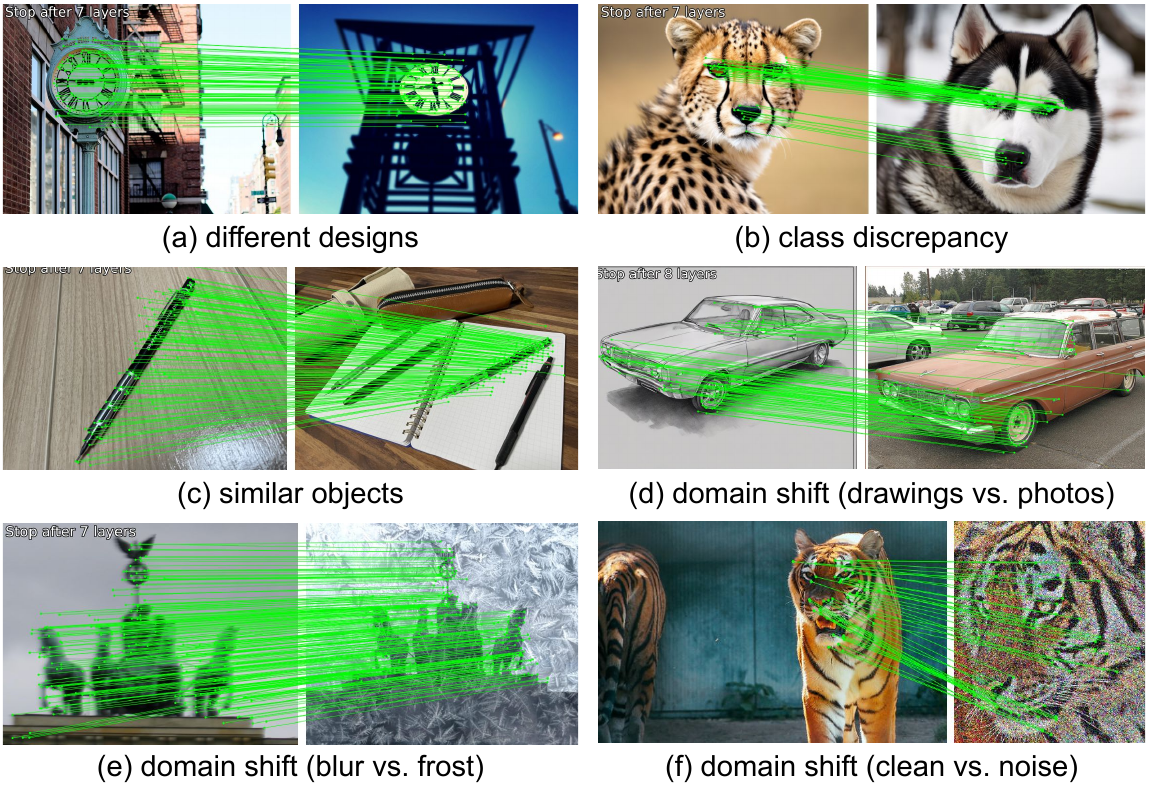}
  \caption{Non-identical object matching between various objects and image styles is achieved by our plug-and-play method with SuperPoint~\cite{SuperPoint} keypoint detector and LightGlue~\cite{LightGlue} matcher. 
  }
  \label{fig:intro_main}
\end{figure}

    In this study, we generalize feature matching and semantic correspondence to \textit{non-identical object matching}, where objects to be matched are not necessarily identical~(see Fig.~\ref{fig:intro_main}). For example, the object pair can be street clocks of different designs or four-legged animals of different species. Further, the non-identicality naturally requires robustness against image corruptions and domain shifts (e.g., photo vs. drawing) because two objects may be captured in different environments and domains. 
    A crucial challenge of achieving non-identical object matching is that we cannot straightforwardly resort to the supervised learning framework because of the ambiguity of non-identicality; there is no ground-truth matching between non-identical objects. 
    
    We propose a plug-and-play module, Semantic Enhancement Weighting (SEW), that extends off-the-shelf matchers to non-identical object matching. The sparse matching pipeline first detects keypoints in a pair of images and feeds their descriptors into a matcher. SEW is inserted before the matcher and weights descriptors using an object detector and Grad-CAM~\cite{GradCAM}. 
    Due to task definition, traditional matching relies on low-level features (e.g., color, edges, texture), which are prone to ambiguity and noise (see Fig.~\ref{fig:task_NIOM}).
    While these features are still vital for fine-grained matching, our high-level feature weighting improves robustness. 
    Further, to handle the case with multiple objects in images, we propose Non-visual Object Pairing, which determines the pairs of similar objects to be matched between images and eliminates mismatching between irrelevant objects.
    
    Our experiments demonstrate that our approach significantly boosts the robustness of sparse matching methods, including LightGlue~\cite{LightGlue} and GlueStick~\cite{GlueStick}, in two tasks: non-identical object matching and robust image matching.
    For the former, we evaluate the matching results with various image pairs (a) of the same class, (b) with class discrepancy, (c) with domain shift, and (d) of the same appearance, and further provide a quantitative evaluation with a new annotation-free metric, \emph{Triangular Matching Consistency} (TMC). 
    While sparse matchers alone yield scattered mismatches, our method achieves fine-grained and consistent alignment, as reflected in higher TMC scores.
    For the latter, we evaluate the robustness of matching on the MegaDepth-1500~\cite{LoFTR} dataset under common corruptions~\cite{CommonCorruption} in terms of the AUC score of standard relative pose estimation.  
    Our method outperforms the state-of-the-art sparse matcher under various types of image corruptions.
    The average AUC of our method even exceeds that of dense matchers, LoFTR~\cite{LoFTR} and Efficient LoFTR~\cite{EfficientLoFTR}, which pursue matching quality at the cost of speed.

    To summarize, our contributions are as follows:
    \begin{itemize}
    \item We tackle non-identical object matching, a novel task of matching similar but not necessarily identical objects in images, even under image corruptions and/or across different domains, which can be considered a generalization of feature matching and semantic correspondence.
    \item We propose Semantic Enhancement Weighting~(SEW), which enhances low-level features in a plug-and-play manner using semantic information and successfully extends various matchers to non-identical object matching. 
    \item Our experiments show that under challenging conditions such as class discrepancy, image corruptions, and domain shifts, our method yields fine-grained and consistent correspondences, validated by TMC and pose estimation.
    \end{itemize}

\section{Related Work}
\label{sec:rel}
\begin{figure}[tb]
  \centering
  \includegraphics[width=\linewidth]{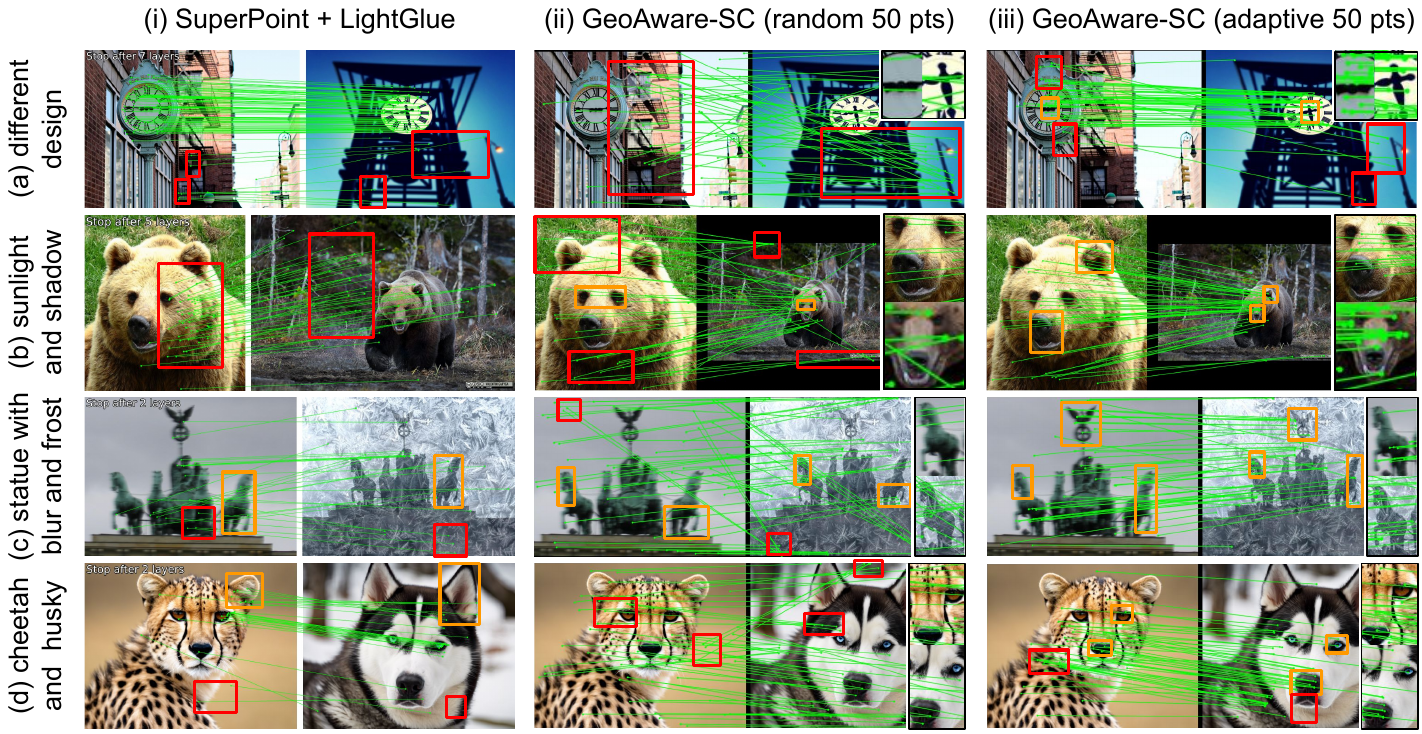}
  \caption{
  State-of-the-art models of sparse matching~(LightGlue~\cite{LightGlue}) and semantic correspondence~(GeoAware-SC~\cite{GeoAwareSC}). 
  The latter requires reference points in one image and finding their correspondence in the target image. To this end, 50 reference points are sampled randomly or based on the GradCAM~\cite{GradCAM} heatmap of YOLOv7~\cite{YOLOv7}~(cf. Sec.~\ref{suppsec:GeoAwareMatchingAlgorithm} in the supplementary material).
  Red boxes show mismatches.
  Orange boxes show close matches, which are correct at the part level, but not at the pixel level. This was typical for Geo-Aware-SC (e.g., matching between the nose tip of the cheetah and the nostrils of the husky in (d)).
  Further, Geo-Aware-SC was about ten times slower than LightGlue.~(See Tab.~\ref{tab:geoaware_ours} in the supplementary material).
  }
  \label{fig:task_NIOM}
\end{figure}
    Image matching~\cite{SuperGlue, LoFTR, LightGlue, DarkFeat} is a fundamental task in computer vision with various applications~\cite{SfM, ORB_SLAM, monoSLAM, self_supervised_keypoint_tracking, denseTracking, TrackingEverything, ObjectPoseEstimation, onepose, DenseMatchSummarization, MATCHA} that involves finding the fine-grained correspondence between two images.
    Many image matching methods have two stages. A keypoint detector first extracts keypoints with their positions and local feature descriptors, and then feature matchers find matches between them.

\medskip\noindent\textbf{Keypoint detectors.}
Classical detectors such as SIFT~\cite{SIFT} and ORB~\cite{ORB} rely on hand-crafted features. 
Recent learning-based methods, including SuperPoint~\cite{SuperPoint} and DISK~\cite{DISK}, achieve higher robustness, with lightweight~\cite{silk} and unsupervised variants~\cite{diffusionkeypoints} also proposed. 

\medskip\noindent\textbf{Feature matchers.}
Feature matchers~\cite{feature_matching_survey, feature_matching_survey2} find a fine-grained correspondence between images. 
The feature matchers are categorized into two types based on the density of matching points: sparse matchers and dense matchers.
Sparse matchers (e.g., SuperGlue~\cite{SuperGlue}, which combines Transformer attention with optimal transport; LightGlue~\cite{LightGlue}, which introduces a hierarchical matching structure for real-time accuracy; GlueStick~\cite{GlueStick}, which exploits line segments as structural features) use limited keypoints for efficiency. 
In contrast, dense matchers (e.g., LoFTR~\cite{LoFTR}, DKM~\cite{DKM}, ROMA~\cite{roma}) pursue matching accuracy rather than speed by matching all pixels. 

\medskip\noindent\textbf{Semantic correspondence models.}
Semantic Correspondence~\cite{SCNet, cats, demystifyingSC, diffhyper} is a task of identifying a semantically similar point in the target image from the given point in the source image, with benchmarks such as SPair-71k~\cite{SPair71k} and PF-PASCAL~\cite{PFPASCAL}. 
Recent models include GeoAware-SC~\cite{GeoAwareSC}, which integrates geometric cues with pre-trained features, and unsupervised methods without annotations~\cite{unsupervisedSC}. 
It is also worth noting that semantic correspondence models run significantly slower than sparse matching methods because feature extraction depends on large generative models, such as  Stable Diffusion~\cite{stable_diffusion}. 
For example, it runs roughly 10 times slower in Fig.~\ref{fig:task_NIOM}.

\medskip
This study addresses non-identical object matching, unifying the fine-grained alignment of feature matching and the high-level matching of semantic correspondence. This makes supervised training technically challenging but also offers unique applications (cf. Sec.~\ref{sub_task_distinction}). 
In another view, we aim for matching that runs in real-time and is robust to object identity variations, corruptions, and domain shifts. 
This covers broader robustness than prior studies~\cite{weaklysemanticmatching, SFD2, dinoloftr}, including DarkFeat~\cite{DarkFeat} and SAM~\cite{sceneaware} for lighting changes.
See Sec.~\ref{secsupp:rel_details} in the supplementary for detailed related work.

\section{Non-Identical Object Matching}
\label{sec:task}
    We introduce non-identical object matching, matching between unnecessarily identical objects in two images. 
    For example, a dog and a cat are not identical but still have similar structures, such as two eyes, four legs, and one tail. The same mass-produced products, such as cars of the same model, have the same appearance but are still not identical (e.g., owned by different people) and can appear in images with largely different backgrounds and lighting conditions.

\subsection{Task formulation}
    We introduce a more formal setup of non-identical object matching. To this end, we first present fine-level image matching~(i.e., dense pixel-wise matching, as opposed to coarse-level image matching tasks such as semantic correspondence) as a general task and then discuss feature matching and non-identical object matching as special cases. 
    \begin{problem}[fine-level image matching]
        Let $\cX$ be the image domain. 
        Given two images $\bm{x}_A,\bm{x}_B \in \cX$, let $D_A = \{(\bm{p}_i^{A}, \bm{d}_i^{A})\}_{i=1}^{n_A}$ be $n_A$ keypoints in image $\bm{x}_A$, where $\bm{p}_i^{A}\in[0,1]^2$ and $\bm{d}_i^{A}\in[0,1]^d$ are its position and descriptor of the $i$-th keypoint.
        We define $D_B = \{(\bm{p}_i^{B}, \bm{d}_i^{B})\}_{i=1}^{n_B}$ for image $\bm{x}_B$ similarly.
        Let $\cL(\,\cdot\,)$ be a (conceptual) matching loss function.
        The task can be formalized as follows. 
        \begin{align}
            \min\quad & \cL(\{m_{kl}\}_{k=1,\ldots, n_A, l=1,\ldots, n_B})\\
            \mathrm{s.t.}\quad & m_{kl} \in [0,1], \forall k \in \{1,\ldots, n_A\}, l \in \{1, \ldots, n_B\} \\
                               & \sum_{l=1}^{n_A} m_{kl}\le 1, \forall k \in \{1,\ldots, n_A\} \\
                               & \sum_{k=1}^{n_B} m_{kl}\le 1, \forall l \in \{1,\ldots, n_B\}.
        \end{align}
    \end{problem}
    The matching coefficient $m_{kl} = 1$ indicates that the $k$-th keypoint in image $\bm{x}_A$ and the $l$-th keypoint in image $\bm{x}_B$ is matched and otherwise $m_{kl} = 0$. Each keypoint in one image has at most one matched point in the other image. This condition is implemented by the last two constraints. 
    
    In the feature matching task, where an identical object is captured from different angles in two images, matching loss $\cL(\{m_{kl}\}_{k,l})$ measures the discrepancy between a camera pose computed the matching $\{m_{kl}\}_{k,l}$ and the ground truth.
    
    Our task, non-identical object matching covers a broader class of matching. Given two images $\bm{x}_A, \bm{x}_B\in\cX$ that contain (potentially non-identical) objects to be matched, image $\bm{x}_I$ can be characterized by several attributes $(o_I, y_I, P_I, D_I)$ for $I\in\{A,B\}$. If $o_A = o_B$, then the objects to be matched are identical. If $o_A\equiv o_B$, they are the same in appearance but not necessarily identical~(e.g., two cars of the same model). If $y_A=y_B$, they belong to the same class (e.g., the car class). If $P_A = P_B$, then they are captured from the same camera pose. If $D_A = D_B$, they are captured in the same domain~(e.g., the drawing domain).
    
    The classical feature matching corresponds to the case with $o_A = o_B$ (consequently, $o_A\equiv o_B$ and $y_A=y_B$), $P_A\ne P_B$, and $D_A=D_B$. In words, it considers identical objects captured from different camera poses in the same domain. In contrast, non-identical object matching encompasses $y_A\ne y_B$ (consequently, $o_A\neq o_B$ and $o_A\not\equiv o_B$), $P_A\ne P_B$, and $D_A\ne D_B$. Note that we practically assume $o_A$ and $o_B$ to be \textit{similar} in some sense. 
    Below, we highlight two important special cases, both assuming $P_A\ne P_B$.
\begin{figure}[tb]
  \centering
  \includegraphics[width=\linewidth]{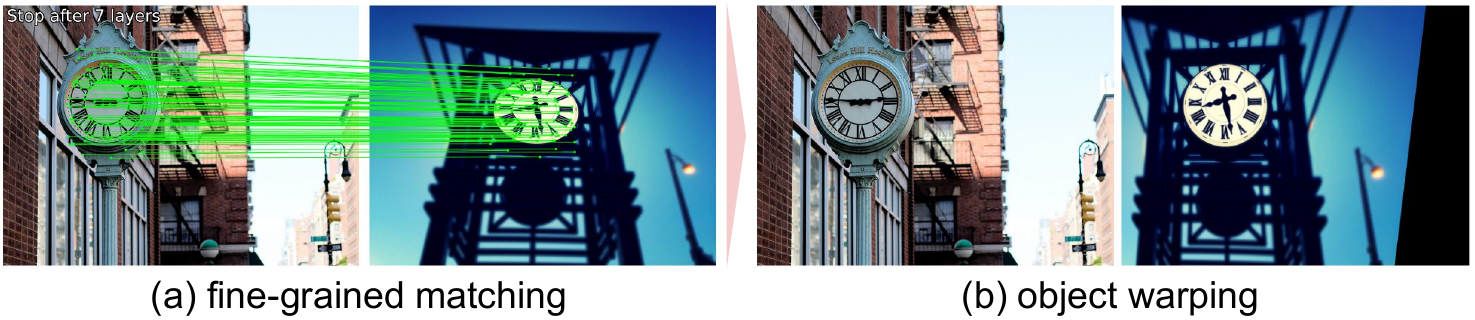}
  \caption{Non-identical object warping.  
  Homography is estimated from fine-grained matching of city clocks with different styles.
  }
  \label{fig:task_warping}
\end{figure}

\medskip\noindent\textbf{Class discrepancy $(y_A \ne y_B)$.}
    Four-legged animals, such as tigers and cats, belong to different classes but have similar structures (two eyes, four legs, and a tail). Despite differences in color, texture, and shape, we humans can still find part-by-part matches. However, it is not trivial to design matching methods to find fine-grained correspondence while being robust to detail differences.

\medskip\noindent\textbf{Domain shift $(o_A \equiv o_B,D_A \ne D_B)$.}
    We humans can know the flower in the painting on the wall depicts that in the garden. This is a robust fine-grained matching of an identical object in different domains. 
    Typical domain shifts include photos vs. drawings and clean vs. corrupted images. For instance, if $\bm{x}_A$ is a photo and $\bm{x}_B$ is a painting, the latter lacks colors, textures, and shadows, causing its descriptors to differ significantly. Without sufficient robustness, this discrepancy leads to missing correspondences.

\subsection{Distinction from prior tasks and applications} \label{sub_task_distinction}
Non-identical object matching relates to feature matching (i.e., fine-level perspective alignment) and semantic correspondence (i.e., coarse-level part correspondences between similar objects). However, achieving fine-grained and semantic alignment of different objects simultaneously is technically non-trivial and has interesting applications.

Non-identical object matching is challenging because it does not readily fit into supervised training. 
There is no available dataset with fine-grained human annotation of matching between distinct objects, and the construction would be laborious.
In contrast, feature matching relies on supervised training with ground-truth homography from 3D reconstruction, while semantic correspondence relies on sparse part-level annotations. Yet its standard metric, PCK~\cite{SPair71k, PFPASCAL}, evaluates only coarse part matches and is therefore inadequate for fine-grained alignment.

The applications of non-identical object matching are unique and cannot be attained by feature matching and semantic correspondence.  
A fine-grained matching with robustness to class discrepancy allows transferring fine-grained landmark annotations. For example, computer vision for animals works on datasets targeting a single animal species, such as facial datasets of dogs~\cite{dogfacial}, cats~\cite{catfacial}, and cattle~\cite{cattlefacial}. Non-identical object matching allows transferring their annotations to other animals. 
Its robustness to domain shift can also compare the sketch or blueprint of a product to a real one, which helps workers assemble parts and build it up. Figure~\ref{fig:task_warping} illustrates a simple example, \textit{non-identical object warping}, that is not covered by prior tasks.

\begin{figure*}[tb]
  \centering
  \includegraphics[width=0.8\linewidth]{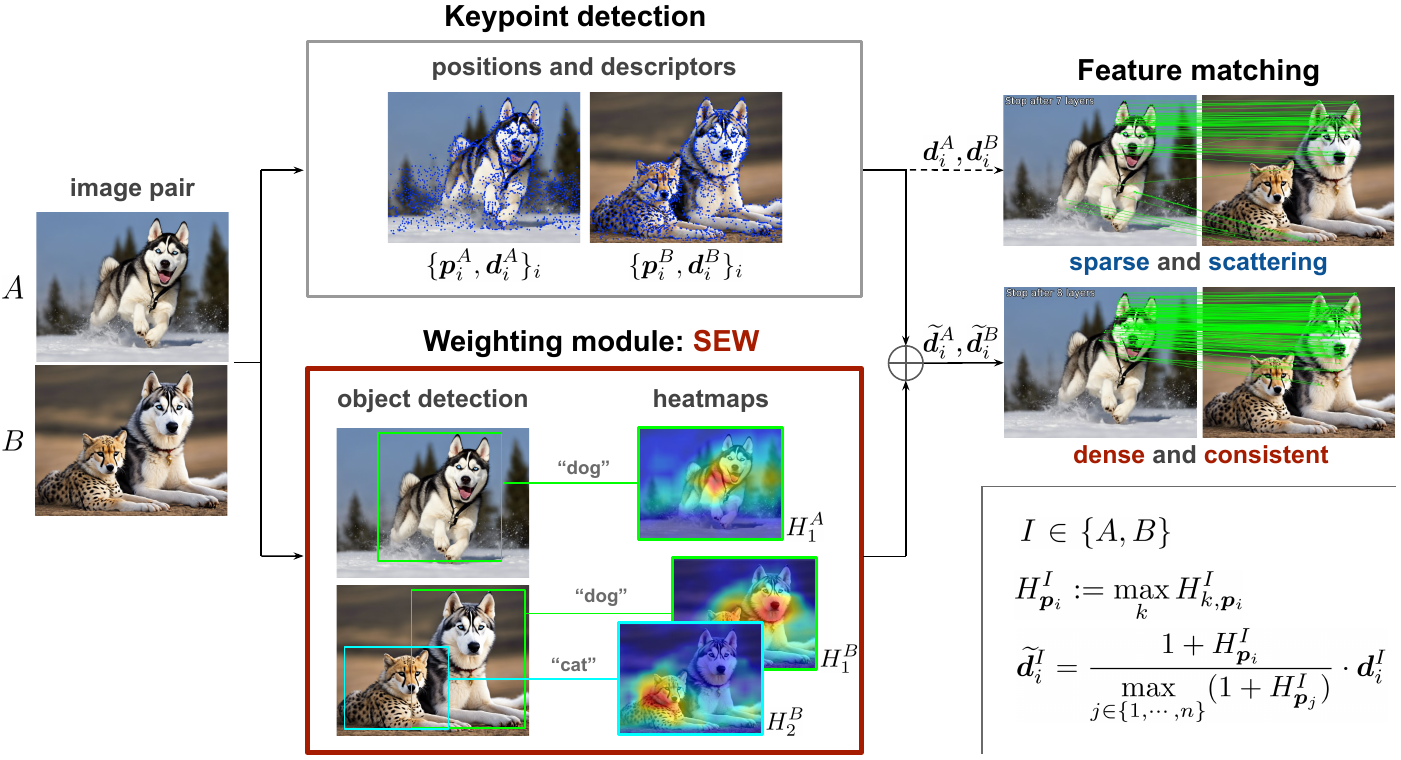}
  \caption{Pipeline of the matching. The keypoint detection and feature matching are done by off-the-shelf models. The proposed plug-and-play module, SEW, computes the heatmap scores of objects and weights the descriptors with this semantic information. 
  }
  \label{fig:method_architecture}
\end{figure*}

\section{Proposed Method}
\label{sec:method}
     We propose a plug-and-play module, Semantic Enhancement Weighting~(SEW), that can be inserted into various sparse matchers, such as SuperGlue~\cite{SuperGlue}, LightGlue~\cite{LightGlue}, and more~\cite{sgmnet, GlueStick}. 
    Sparse matchers consist of a pair of a keypoint detector and a feature matcher. Our module processes the keypoints before feeding them to the feature matcher.
    Particularly, given a pair of images $\bm{x}_A,\bm{x}_B \in \cX$, The keypoint detector outputs collections of keypoints $D_A = \{(\bm{p}_i^A, \bm{d}_i^A)\}_i, D_B = \{(\bm{p}_i^B, \bm{d}_i^B)\}_i$. The proposed module then computes weighted descriptors $\widetilde{\bm{d}}_i^A, \widetilde{\bm{d}}_i^B$ for each $\bm{d}_i^A,\bm{d}_i^B$, respectively. Lastly, the feature matcher performs matching using $\widetilde{D}_A = \{(\bm{p}_i^A, \widetilde{\bm{d}}_i^A)\}_i, \widetilde{D}_B = \{(\bm{p}_i^B, \widetilde{\bm{d}}_i^B)\}_i$.
    
    As will be shown in Sec.~\ref{sec:experiments}, this simple pipeline works surprisingly well and adapts pretrained sparse matching models without any training, addressing the technical challenge in non-identical object matching discussed above.

\subsection{Semantic Enhancement Weighting}\label{sec:methodweighting}
    The proposed weighting module uses an object detector and a visual explanation model. 
    This study used YOLOv7~\cite{YOLOv7} and Grad-CAM~\cite{GradCAM}. 
    Let $\bm{x} \in \cX$ be an image, and let $D = \{(\bm{p}_i, \bm{d}_i)\}_{i=1}^n$ be the collection of $n$ keypoints of $\bm{x}$. 

    First, the object detector is applied to $\bm{x}$ and detects the objects with their classes. For each of the detected objects (say, the $k$-th object), the visual explanation model gives pixel-wise class-activation scores or a heatmap $H_k$. Let $H_{k,\bm{p}_i} \in [0,1]$ be the heatmap score of $H_k$ at $\bm{p}_i$. Then, the scores at $\bm{p}_i$ are aggregated as $H_{\bm{p}_i} := \max_k H_{k,{\bm{p}_i}}$.\footnote{The max operation outperformed other candidates including mean, sigmoid, and softmax operations in our preliminary experiments.}
    Finally, the weighted descriptor $\tilde{\bm{d}}_i$ is computed by 
    \begin{align} \label{eq:weight}
        \widetilde{\bm{d}}_i = \frac{1+H_{\bm{p}_i}}{\underset{j \in \{ 1, \cdots , n \}}{\max}(1+H_{\bm{p}_j})} \cdot \bm{d}_i \in [0.5, 1]^d.
    \end{align}
    We empirically found that the addition of base value 1 and the normalization by the maximum value is important. For example, the simplest approach $\widetilde{\bm{d}}_i = H_{\bm{p}_i}\bm{d}_i \in [0, 1]^d$ ends up with a scarcity of matching because of the heatmap scores on the keypoints on background tend to be too low, i.e., $H_{\bm{p}_i} \approx 0$. Although such keypoints will not likely appear in the matching results, they seem to help the keypoints on objects to be matched. 
    If one instead uses $\widetilde{\bm{d}}_i = (1+H_{\bm{p}_i})\bm{d}_i \in [1, 2]^d$, this also affects negatively. We consider that this is because of the softmax operations used in the subsequent feature matching. We next discuss how the weighting of descriptors impacts the matching.

\subsection{Effect of weights in feature matcher}
    We now illustrate how our weighting module affects the subsequent feature matching. We take LightGlue~\cite{LightGlue} as an example, but a similar discussion holds for other feature matchers.
    For simplicity, let $\alpha_{i}$ be the weight in \cref{eq:weight}, i.e., $\widetilde{\bm{d}}_i = \alpha_i \bm{d}_i$. The matcher takes as input the keypoints $\widetilde{D}_A = \{(\bm{p}_i^A, \widetilde{\bm{d}}_i^A)\}_i, \widetilde{D}_B = \{(\bm{p}_i^B, \widetilde{\bm{d}}_i^B)\}_i$. The vectors $\widetilde{\bm{d}}_i^A = \alpha_i^A \bm{d}_i^A, \widetilde{\bm{d}}_i^B = \alpha_i^B \bm{d}_i^B$ are the $i$-th weighted descriptors of image $\bm{x}_A, \bm{x}_B$, respectively. Below, symbols with and without $\widetilde{\,\cdot\,}$ present those with and without weighting. 
    
    In the feature matcher, the descriptors are first converted to key and query vectors using matrices $W_{q}, W_{k} \in \bR^{d\times d}$. 
    \begin{align} \label{eq:querykey}
        \widetilde{\bm{q}}_i^I &= W_q \widetilde{\bm{d}}_i^I = \alpha_iW_q\bm{d}_i^I = \alpha_i\bm{q}_i^I \\
        \widetilde{\bm{k}}_i^I &= W_k \widetilde{\bm{d}}_i^I = \alpha_iW_k\bm{d}_i^I = \alpha_i\bm{k}_i^I,
    \end{align}
    where $I \in \{A, B\}$. Then, the self-attention score $a_{ij}^{I}$ between $i$-th and $j$-th keypoints of image $\bm{x}_I$ is calculated as
    \begin{align} \label{eq:self}
        \widetilde{a}_{i j}^I&=\widetilde{\bm{q}}_{i}^{I\top} \mathcal{R}\left(\bm{p}_{j}^I-\bm{p}_{i}^I\right) \widetilde{\bm{k}}_{j}^I \\ 
        &= (\alpha_i^I\alpha_j^I)\cdot\qty( \bm{q}_{i}^{I\top} \mathcal{R}\left(\bm{p}_{j}^I-\bm{p}_{i}^I\right) \bm{k}_{j}^I) \\
        &= (\alpha_i^I\alpha_j^I)\cdot a_{ij}^I,
    \end{align}
    where $\mathcal{R}: \bR^{2} \to \bR^{d \times d}$ denotes the rotary-encoding~\cite{RoFormer}. Therefore, our weighting by heatmap scores leads to the weighting of attention scores. The self-attention scores reflect the internal relationship of objects and background, thereby extracting better feature representations in the subsequent steps of the feature matcher. Roughly speaking, noting that $\alpha_i^I \in [0.5, 1.0]$ for all $i$ and $I \in \{A, B\}$, we have $\alpha_i^I\alpha_j^I \approx 1$ between the keypoints on objects, $\alpha_i^I\alpha_j^I \approx 0.5$ between one on object and the other on background, and $\alpha_i^I\alpha_j^I \approx 0.25$ for between those on the background. This is reasonable because we want nice feature representations for object keypoints for nice matching. Nevertheless, such nice representations should be obtained by taking into account the background conditions, e.g., to compensate for the overall lighting conditions. 
    If one uses $\alpha_i^I = H_{\bm{p}_i}$ as discussed earlier, i.e., $H_{\bm{p}_i}\approx 0$ for background keypoints, the feature representations cannot take into account the background.

    Similarly, our weighting affects the cross-attention score $a_{i j}^{A B}$ between the $i$-th keypoint of image $\bm{x}_A$ and the $j$-th keypoint of image $\bm{x}_B$, computed from the key vectors.
    \begin{align} \label{eq:cross}
        \widetilde{a}_{i j}^{A B} = \widetilde{\bm{k}}_{i}^{A \top} \widetilde{\bm{k}}_{j}^{B} = (\alpha^A_i\alpha^B_j)\cdot a_{ij}^{AB}.
    \end{align}
    As discussed earlier, $\alpha_i^A\alpha_j^B$ is large between keypoints of the objects, encouraging matching between objects.

\subsection{Non-visual Object Pairing}
\label{sec:class_similarity_adapter}
\begin{figure}[tb]
  \centering
      \includegraphics[width=\linewidth]{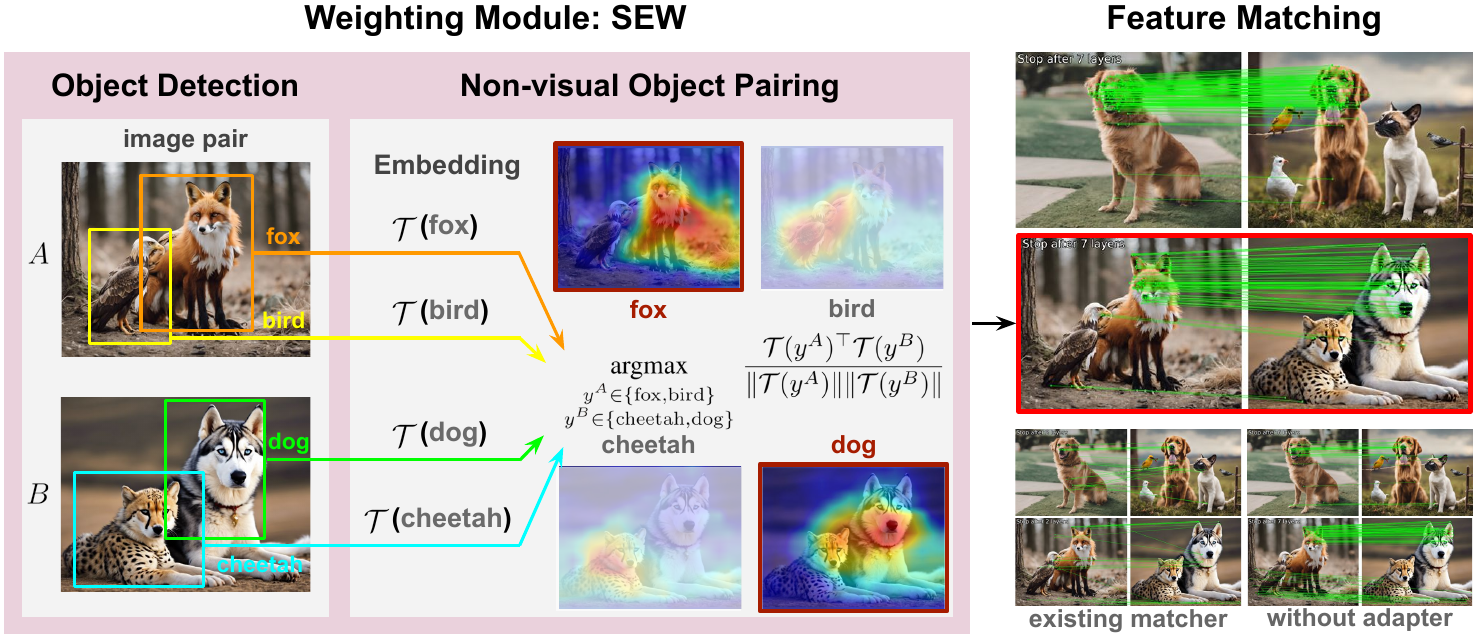}
      \caption{Non-visual Object Pairing converts class labels from an object detector into embeddings using the CLIP~\cite{CLIP} text encoder and compares them with cosine similarity to identify semantically similar object pairs.
      The GradCAM~\cite{GradCAM} heatmap of object detector YOLOv7~\cite{YOLOv7} generated for the chosen object pair is used for our weighting method, enhancing matching with multiple objects. 
      }
  \label{fig:method_adapter}
\end{figure}

We next propose Non-visual Object Pairing for the case where images contain multiple objects. 
In such a case, it is essential to determine which object in one image should be matched with another object in the other image, and the weighting module should exploit the heatmaps of them and exclude those of irrelevant objects. 
Otherwise, the local spurious similarity of objects causes scattered and inconsistent matching; see ~\cref{fig:method_adapter}~(bottom right), where a nose of a dog and an ear of a cat match.

A pair of \textit{similar} objects for matching can be determined using visual or non-visual features. For the former, we can use the feature vectors extracted from an intermediate layer of the object detector. This can be done simultaneously during the heatmap computation process. For the latter, text embedding (e.g., by the CLIP text encoder) of object labels may be a reasonable choice. The text embedding vectors can be pre-computed for all the classes if one pursues fast matching.
The visual and non-visual features are complementary, and the choice should depend on the use case.
For example, given (\textit{lesser panda}, \textit{raccoon}, \textit{giant panda}),  the visually similar pair is  (\textit{lesser panda}, \textit{raccoon}), while  (\textit{lesser panda}, \textit{giant panda}) is more conceptually similar.

We here provide concrete procedures in the Non-visual Object Pairing based on non-visual features. Suppose that an image $\bm{x}_A$ contains $n_A$ objects, and let $Y_A = \{y_i^A\}_{i=1}^{n_A}$ be their class labels. We define $Y_B = \{y_i^B\}_{i=1}^{n_B}$ similarly for the other image $\bm{x}_B$.
Let $\mathcal{T}$ be a pretrained text encoder that maps a label $y$ to an embedding vector $\mathcal{T}(y)$. Particularly, we used the CLIP text encoder~\cite{CLIP}.
The adapter determines the object pair $(i, j)$ to be matched based on the cosine similarity between the embedding vectors as follows.
\begin{align}
    (y_i^A, y_j^B) = \underset{
        y^A \in Y_A,~y^B \in Y_B
    }{\text{argmax}} \; \frac{\mathcal{T}(y^A)^\top \mathcal{T}(y^B)}{\|\mathcal{T}(y^A)\| \|\mathcal{T}(y^B)\|},
\end{align}
where $\norm{\,\cdot\,}$ denotes the $L_2$ norm of vector.
Then, the weighting module only uses the Grad-CAM heatmaps of the $i$-th object in $\bm{x}_A$ and the $j$-th object in $\bm{x}_B$. The process is almost the same when visual features are used; the text embedding vectors are replaced by the feature vectors from the intermediate layer of the object detector.
Finally, the weighting module uses the corresponding heatmaps for weighting the descriptors.
Figure~\ref{fig:method_adapter} shows this pipeline.

\section{Experiments}\label{sec:experiments}
    In this section, we demonstrate the effectiveness of the proposed method in non-identical object matching tasks.  
    Section~\ref{sub:exp_NIOM} presents various qualitative comparisons from state-of-the-art sparse feature matching methods. 
    Section~\ref{sub:exp_TMC} then introduces \emph{Triangular Matching Consistency} (TMC), a new annotation-free metric that allows us to quantitatively evaluate non-identical object matching itself. 
    Section~\ref{sub:exp_Pose} presents quantitative evaluations. In the latter experiment, as there are no available datasets providing fine-grained matching between non-identical objects, we examine the matching results between identical objects under 15 types of corruptions, which measures the robustness to domain shifts. 
    We emphasize that while semantic correspondence is a conceptually similar task to ours, it does not target fine-grained matching, and thus, our experiment setup~(i.e., datasets, baseline methods, and metrics) is based on feature matching literature. 
    
\begin{figure}[tb]
  \centering
  \includegraphics[width=\linewidth]{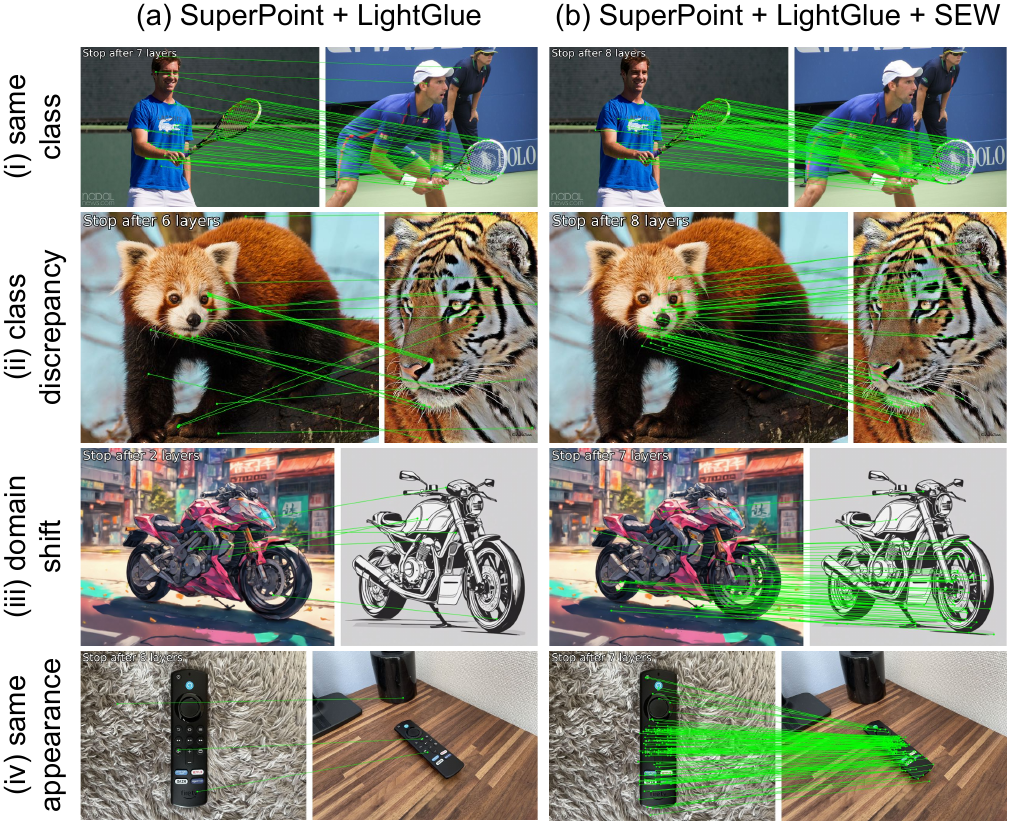}
  \caption{Typical categories of non-identical object matching. (a) A combination of SuperPoint~\cite{SuperPoint} and LightGlue~\cite{LightGlue} only finds a small number of correspondences, and many of them are incorrect. (b) Our method SEW significantly improves the matching.
  }
  \label{fig:exp_NIOM}
\end{figure}

\medskip\noindent\textbf{Setup.} 
The sparse matching pipeline involves a keypoint detector and a feature matcher.
For the former, we adopted pre-trained SuperPoint~\cite{SuperPoint}, extracting 2,048 keypoints. For the latter, the extracted keypoints are matched using either pretrained SGMNet~\cite{sgmnet}, GlueStick~\cite{GlueStick}, and the state-of-the-art method, LightGlue~\cite{LightGlue}, with and without the proposed method. We additionally tested a detector-free dense matcher, LoFTR~\cite{LoFTR} and Efficient LoFTR~\cite{EfficientLoFTR}. We downloaded the pretrained models from their official repositories (details provided in the supplementary material).
All experiments used NVIDIA GeForce RTX 3090 GPU and 24GB memory.

\subsection{Matching Non-Identical Objects} \label{sub:exp_NIOM}
Here, we perform the matching of non-identical objects and evaluate them qualitatively. 

\medskip\noindent\textbf{Datasets.}
    We collected various images with objects from the COCO~\cite{cocodataset} and ImageNet~\cite{imagenet} datasets. To prepare the drawing version of several images, we used the DreamStudio\footnote{\url{https://beta.dreamstudio.ai/generate}}, which is backended by the Stable Diffusion~\cite{stable_diffusion}.

\medskip\noindent\textbf{Robust matching across non-identical objects.}
We examined four cases: (i) same class (i.e., $y_A = y_B, D_A = D_B$), (ii) class discrepancy (i.e., $y_A \ne y_B, D_A = D_B$), (iii) domain shift (i.e., $o_A \equiv o_B$, $D_A\ne D_B$), and (iv) same appearance (i.e., $o_A = o_B$). For all the cases, we have a perspective difference (i.e., $P_A \ne P_B$). Due to the page number restriction, we only present a limited number of matching examples here, but more results can be found in Sec.~\ref{suppsec:moreNIOM} in the supplementary material. 
Figure~\ref{fig:exp_NIOM} shows the matching results with and without the proposed method in the four cases. 
As shown in \cref{fig:exp_NIOM}~(a), SuperPoint and LightGlue alone fail to distinguish objects from the background or match corresponding animal structures. 
In contrast, \cref{fig:exp_NIOM}~(b) shows that the proposed method significantly reduces unreasonable matches (e.g., human and racket, animal eye and mouth) while increasing correct part-to-part matchings.
Particularly, it (i) increased racket matches while reducing mismatches between people, (ii) aligned corresponding animal parts, (iii) matched front wheels despite domain gaps, and (iv) significantly increased matches despite background and angle variations.
Notably, structural features, such as eyes and noses, are successfully matched between different classes.
We consider that the heatmaps of these animals gave similar importance to the eyes or mouth, thus increasing the number of matches between the same parts of the body.
In addition, the results for the same class and appearance showed a decrease in mismatches with the background, which means that the background and foreground had relatively lower similarity.

\begin{figure}[tb]
  \centering
  \includegraphics[width=\linewidth]{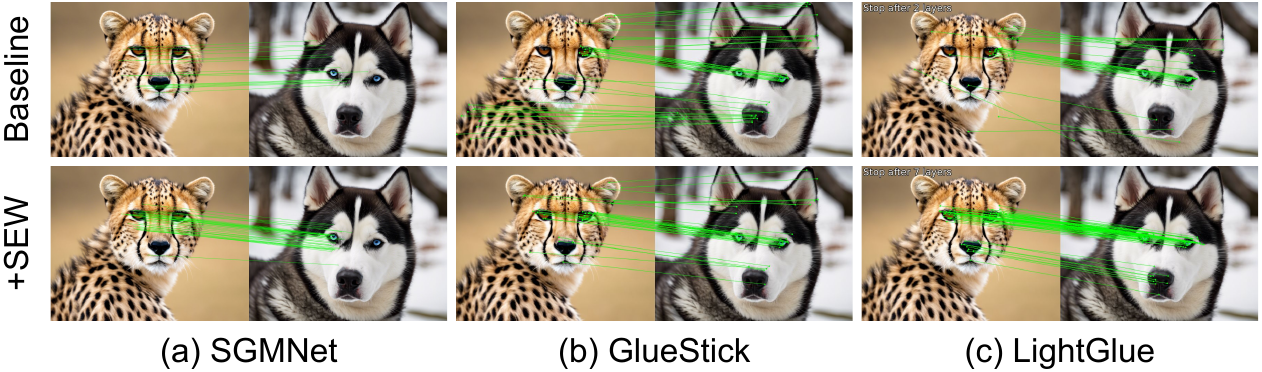}
  \caption{
  The results with different matchers~\cite{sgmnet, GlueStick, LightGlue}. SuperPoint~\cite{SuperPoint} is used for the keypoint detection. The introduction of SEW increases the matching of eyes and noses between the cheetah and husky and decreases mismatching in the background.
  }
  \label{fig:exp_matchers}
\end{figure}

\medskip\noindent\textbf{Different matchers.}
Figure~\ref{fig:exp_matchers} presents the results with other matchers, showing that our method boosts all of them to match corresponding parts between different animals.

\subsection{Triangular Matching Consistency}
\label{sub:exp_TMC}
Direct quantitative evaluation of non-identical object matching is difficult, as fine-grained ground-truth correspondences between non-identical objects are inherently ambiguous. 
We therefore introduce \emph{Triangular Matching Consistency} (TMC), an annotation-free metric. 
TMC measures the agreement between direct matching $(A,C)$ and one-hop matching via $(A,B)$ and $(B,C)$, where $A,B,C$ are semantically related images. 
By selecting $(A,C)$ to be highly similar, their direct matches provide a reliable reference, and a robust our task method should approximate them even through the one-hop path. 
TMC reports error magnitude (RMSE), accuracy (PCK), and coverage (Recall). See Sec.~\ref{supp_TMC_details} and Fig.~\ref{fig:TMC_overview} for the details.

\medskip\noindent\textbf{Dataset.}
We build an image set of 1,500 triplets for TMC. 
We use animal images from the Wild category of the Animal Faces dataset~\cite{animalfaces}.
This dataset contains diverse animal face images, including dogs, cats, lions, tigers, and more.
For each image $A_i$, we generate a new view $C_i$ of the same object by applying a 0.9× pinch-out geometric transformation.
Then, another species image $A_j$ ($j \ne i$) is used as $B$, forming $(A,B,C)=(A_i,A_j,C_i)$. 
Here, $(A,C)$ are identical-instance pairs with viewpoint change, while $B$ provides a semantically similar but different object. 
This design allows us to use $(A,C)$ as a reliable reference while introducing non-identical variation through $B$, which is essential for evaluating our new task. 

\medskip\noindent\textbf{Results.}
Table~\ref{tab:experiment-tmc-wild} summarizes the evaluation.
RMSE measures geometric error, and our module reduces it to about half of each baseline. It even outperforms dense detector-free methods.
PCK measures the proportion of matches within a tolerance. It improves at all thresholds, with the largest gains at broader ones and clear benefits even under strict settings. This shows that our method enhances consistent matching between corresponding object parts.
Recall measures how well one-hop correspondences cover the $A{\leftrightarrow}C$ baseline, and it also increases with our module.
Across all three metrics, it indicates that our method enhances the matching consistency of non-identical objects.

\setlength{\tabcolsep}{2pt}
\begin{table}[t]
    \centering
    \footnotesize
    \caption{Triangular Matching Consistency (TMC) results for the wild category in the Animal Faces dataset~\cite{animalfaces}. 
    Sparse matchers (LightGlue (LG), GlueStick (GS)) with SuperPoint; dense matchers (LoFTR, ELoFTR) are shown for reference. 
    Metrics include RMSE (lower is better), PCK, and Recall at distance thresholds of 0.01, 0.05, and 0.10 of the image size. Our method SEW shows large improvements across all metrics. See Sec.~\ref{secsupp:moreTMCresults} for details.}
    \label{tab:experiment-tmc-wild}
    \begin{tabular}{llcccccccc}
        \toprule
        \multicolumn{2}{c}{\multirow{2}{*}{Matching Method}} &
        \multirow{2}{*}{RMSE$\downarrow$} &
        \multicolumn{3}{c}{PCK$\uparrow$} &
        \multicolumn{3}{c}{Recall$\uparrow$} \\
        \multicolumn{2}{c}{} & & 0.01 & 0.05 & 0.10 & 0.01 & 0.05 & 0.10 \\ \midrule

        \multirow{4}{*}{\shortstack[l]{Sparse Matcher \\ (with SuperPoint)}}
            & LG
                & \multicolumn{1}{c|}{0.14}
                & 54.0 & 68.9 & \multicolumn{1}{c|}{68.2}
                & 70.2 & 76.9 & \multicolumn{1}{c}{81.4} \\ 

            & LG+SEW
                & \multicolumn{1}{c|}{\textbf{0.07}}
                & \textbf{73.8} & \textbf{87.8} & \multicolumn{1}{c|}{\textbf{92.6}}
                & \textbf{73.8} & \textbf{87.7} & \multicolumn{1}{c}{\textbf{92.9}} \\ \cmidrule{2-9}

            & GS
                & \multicolumn{1}{c|}{0.16}
                & 49.0 & 62.1 & \multicolumn{1}{c|}{62.3}
                & 63.8 & 71.3 & \multicolumn{1}{c}{73.0} \\

            & GS+SEW
                & \multicolumn{1}{c|}{0.08}
                & 68.8 & 81.0 & \multicolumn{1}{c|}{86.6}
                & 67.3 & 82.1 & \multicolumn{1}{c}{86.3} \\ \midrule

        \multirow{2}{*}{\shortstack[l]{Dense Matcher \\ (detector-free)}}
            & LoFTR
                & \multicolumn{1}{c|}{0.19}
                & 52.7 & 61.9 & \multicolumn{1}{c|}{66.8}
                & 57.3 & 62.1 & \multicolumn{1}{c}{65.7} \\

            & ELoFTR
                & \multicolumn{1}{c|}{0.14}
                & 64.5 & 73.7 & \multicolumn{1}{c|}{76.7}
                & 65.9 & 73.3 & \multicolumn{1}{c}{76.9} \\
        \bottomrule
    \end{tabular}
\end{table}

\setlength{\tabcolsep}{5pt}
\begin{table*}[t]
    \centering
    \footnotesize
    \caption{The pose accuracy (AUC) at the maximum angular error of $20\tcdegree$ of the relative pose estimation from image pairs MegaDepth-1500~\cite{LoFTR} under common corruptions~\cite{CommonCorruption}. 
    The left side shows both images corrupted with the same type of corruption, while the right side shows only one of the input images corrupted and the other as a clean image.
    Our method SEW largely improves LightGlue (LG)~\cite{LightGlue} and GlueStick (GS)~\cite{GlueStick} for most categories and the average AUC. 
    The dense matchers, LoFTR~\cite{LoFTR} and Efficient LoFTR (ELoFTR)~\cite{EfficientLoFTR}, are shown for reference.
    Our method improves the robustness of sparse matchers with a slight reduction in clean accuracy and runtime.
    }
    \label{tab:experiment-cc-and-clean}
    \begin{tabular}{lcc|cc|cc||cc|cc|cc}
        \toprule
        \multirow{3.5}{*}{Common Corruptions} & 
        \multicolumn{6}{c}{AUC@$20\tcdegree$ with pairs of \textbf{corrupted} images} & 
        \multicolumn{6}{c}{AUC@$20\tcdegree$ with pairs of \textbf{clean and corrupted} images} \\ \cmidrule{2-7} \cmidrule{8-13}
        & 
        \multicolumn{4}{c}{\textit{keypoint detector : SuperPoint}} &
        \multicolumn{2}{c}{\textit{dense matcher}} &
        \multicolumn{4}{c}{\textit{keypoint detector : SuperPoint}} &
        \multicolumn{2}{c}{\textit{dense matcher}} \\
        &
        \multicolumn{1}{c}{LG} &
        \multicolumn{1}{c}{LG+SEW} &
        \multicolumn{1}{c}{GS} &
        \multicolumn{1}{c}{GS+SEW} &
        \multicolumn{1}{c}{LoFTR} &
        \multicolumn{1}{c}{ELoFTR} &
        \multicolumn{1}{c}{LG} &
        \multicolumn{1}{c}{LG+SEW} &
        \multicolumn{1}{c}{GS} &
        \multicolumn{1}{c}{GS+SEW} &
        \multicolumn{1}{c}{LoFTR} &
        \multicolumn{1}{c}{ELoFTR} \\ \midrule
        None (Clean)         & \textbf{80.61}  & 78.42 & \textbf{78.04} & 75.08 & 80.93 & 83.48 & \textbf{80.61}  & 78.42 & \textbf{78.04} & 75.08 & 80.93 & 83.48 \\ 
        \midrule
        Gaussian Noise       & 43.09  & \textbf{53.27}  & 45.67  & \textbf{52.24}  & 34.97  & 35.71   & 27.54  & \textbf{41.01}  & 41.93  & \textbf{44.98}  & 33.02  & 36.56   \\
        Shot Noise           & 43.41  & \textbf{53.82}  & 46.10  & \textbf{51.72}  & 40.00  & 40.36   & 32.10  & \textbf{42.35}  & 41.75  & \textbf{43.82}  & 36.53  & 39.52   \\
        Impulse Noise        & 44.98  & \textbf{50.11}  & 44.91  & \textbf{50.05}  & 37.73  & 39.89   & 35.95  & \textbf{43.67}  & 40.97  & \textbf{42.91}  & 34.92  & 38.63   \\
        Defocus Blur         & 32.69  & \textbf{48.35}  & 45.58  & \textbf{49.23}  & 57.22  & 54.97   & 18.25  & \textbf{32.13}  & 23.16  & \textbf{30.10}  & 49.40  & 41.97   \\
        Frosted Glass Blur   & 33.37  & \textbf{47.88}  & 47.40  & \textbf{49.94}  & 54.48  & 47.09   & 34.25  & \textbf{44.80}  & 42.96  & \textbf{46.21}  & 54.53  & 47.52   \\
        Motion Blur          & 42.12  & \textbf{53.63}  & 44.76  & \textbf{51.84}  & 55.60  & 48.12   & 50.07  & \textbf{53.10}  & 44.27  & \textbf{52.69}  & 52.81  & 45.10   \\
        Zoom Blur            & 24.40  & \textbf{31.21}  & 22.87  & \textbf{30.97}  & 23.08  & 22.54   & 34.67  & \textbf{40.04}  & 31.18  & \textbf{37.11}  & 31.09  & 28.49   \\
        Snow                 & \textbf{31.51}  & 30.33  & \textbf{25.26}  & 23.21  & 31.85  & 37.89   & 56.24  & \textbf{58.15}  & 47.22  & \textbf{48.53}  & 49.15  & 54.56   \\
        Frost                & \textbf{32.24}  & 31.71  & \textbf{26.39}  & 24.22  & 19.23  & 25.49   & 62.20  & \textbf{64.87}  & 54.34  & \textbf{57.26}  & 43.17  & 45.18   \\
        Fog                  & 70.99  & \textbf{73.49}  & 71.19  & \textbf{73.89}  & 67.10  & 67.74   & 76.96  & \textbf{78.40}  & \textbf{75.18}  & 74.87  & 74.06  & 75.36   \\
        Brightness           & \textbf{75.48}  & 75.08  & \textbf{73.03}  & 70.84  & 75.45  & 79.83   & \textbf{77.14}  & 76.98  & \textbf{76.78}  & 75.06  & 78.41  & 81.26   \\
        Contrast             & \textbf{39.50}  & 38.47  & 39.47  & \textbf{42.61}  & 53.22  & 59.78   & 43.09  & \textbf{45.17}  & 40.44  & \textbf{43.53}  & 39.66  & 52.76   \\
        Elastic Transform    & 54.78  & \textbf{66.21}  & 60.37  & \textbf{64.36}  & 58.56  & 60.50   & 64.93  & \textbf{68.34}  & 68.02  & \textbf{69.92}  & 68.95  & 71.49   \\
        Pixelate             & 67.94  & \textbf{68.24}  & \textbf{65.01}  & 64.81  & 74.77  & 77.27   & 66.80  & \textbf{68.43}  & \textbf{67.16}  & 65.46  & 76.21  & 78.45   \\
        JPEG Compression     & 29.54  & \textbf{36.97}  & 39.66  & \textbf{44.70}  & 59.05  & 59.80   & 47.41  & \textbf{53.82}  & \textbf{60.39}  & 60.12  & 68.14  & 71.70   \\  \midrule
        Average              & 44.40  & \textbf{50.58}  & 46.51  & \textbf{49.64}  & 49.49  & 50.47   & 48.51  & \textbf{54.08}  & 50.38  & \textbf{52.84}  & 52.67  & 53.90   \\ 
        Time [ms/pair]   & 46.67   & 68.43 & 105.72 & 128.18 & 312.54 & 191.17 & 45.98   & 68.37 & 105.34 & 127.86 & 305.01 & 192.53 \\  \bottomrule
    \end{tabular}
\end{table*}

\subsection{Robust Image Matching} \label{sub:exp_Pose}
    We next quantitatively evaluate the proposed method on a standard image matching task with common corruptions. 
    To our knowledge, image matching methods have only been evaluated clean images in literature, except for a few specialized to low-lighting~\cite{DarkFeat} and thermal imaging~\cite{xoftr}. 
    We address two cases: (i) both input images are corrupted, and (ii) only one of them is corrupted. 
    The former shows robustness to corruptions, while the latter shows it to domain shifts (i.e., clean image domain vs. corrupted image domain).

\medskip\noindent\textbf{Dataset.}
    We used the MegaDepth-1500~\cite{LoFTR} test set because it contains many objects, making it suitable for a fair evaluation of the non-identical object matching task. 
    This is not the case with other standard datasets~\cite{Aachen_RobotCar_Seasons_dataset, scannet_dataset, hpatches_dataset, yfcc100m_dataset, ETH_dataset}.
    MegaDepth~\cite{MegaDepth} dataset consists of 3D reconstructed data from one million images for two popular phototourism destinations, with camera poses and depth maps computed via COLMAP~\cite{SfM} and refined as ground truth.
    Common Corruptions~\cite{CommonCorruption} simulates the 15 types of corruptions that occur when images are captured.
    The severity level of corruption is set to 5, which is the most challenging condition.
    
\medskip\noindent\textbf{Metric.}
    We follow the standard evaluation of image matching based on relative pose estimation~(e.g., see~\cite{LightGlue}).
    From the matching results, the relative pose between two images is computed with RANSAC~\cite{RANSAC} as follows.
    First, an essential matrix representing the positional transformation between the images is calculated, which is then decomposed into a rotation and a translation.
    The camera pose error is computed by the maximum angular error, and the area under the cumulative error curve (AUC) at $20\tcdegree$ is calculated.

\medskip\noindent\textbf{Robustness against common corruptions.} \label{subsub_cc}
    The left side of~\cref{tab:experiment-cc-and-clean} shows the AUC of matching when input images are both corrupted.  The results of the dense matching methods (i.e., LoFTR and Efficient LoFTR) are given for reference. As the results show, the proposed method makes the LightGlue and GlueStick robust against most types of common corruptions~(roughly 5\,\% to 10\,\% increase) with a slight increase in runtime. The average AUC even exceeds that of Efficient LoFTR.
    The matching methods benefit from the proposed method, particularly when images are corrupted by noise and blur. We observe AUC decrease in a few types of corruptions, such as Snow and Frost, by the introduction of the proposed method. These corruptions introduce edgy geometric patterns, and the matcher finds correspondence between them.
    See Sec.~\ref{subsec:detailsCC} in the supplementary material for more detailed comparisons.

\medskip\noindent\textbf{Robustness against environmental changes.} 
    As the objects to be matched are non-identical, they can appear in largely different environments. The right side of~\cref{tab:experiment-cc-and-clean} shows the AUC when only one of the input images is corrupted and the other is clean. 
    As in the left side of~\cref{tab:experiment-cc-and-clean}, the proposed method boosts the LightGlue and GlueStick for most cases and the average case.
    Interestingly, the results are better for almost all weather categories.
    Only the brightness category shows a decrease in accuracy, but the drop is slight.
    Weather categories such as Snow and Frost add strong geometry to the overall image, causing mismatches as occlusions.
    Heatmap weighting is considered to structurally complement the areas hidden by occlusion.
    Overall, the proposed method enhances robustness to environmental changes, with only a slight drop in precision on clean images.
    See Sec.~\ref{subsec:detailsCleanCommon} in the supplementary for details.

\section{Conclusion}
    We addressed a novel task of matching non-identical objects, which are only semantically similar, not necessarily identical.
    Due to the non-identicality, two objects may be captured in different environments and domains. 
    Thus, robust matching under image corruptions and domain shifts is also encompassed.
    Our method exploits object detection and visual explanation as high-level features, thereby focusing matching on objects.
    The experiments showed the effectiveness of our method in non-identical object matching using TMC, a new annotation-free metric, and relative pose estimation under various image corruptions.

\section*{Acknowledgement}
Kazuhiko Kawamoto and Hiroshi Kera were supported by JSPS KAKENHI Grant Numbers JP23K24914 and JP22K17962, respectively. They were also supported by Ricoh Company, Ltd.

{
\small
\bibliographystyle{ieeenat_fullname}

}

\newpage
\clearpage
\appendix
\setcounter{page}{1}
\setcounter{figure}{0}
\setcounter{table}{0}
\maketitlesupplementary
\renewcommand{\thesection}{\Alph{section}}
\renewcommand{\thefigure}{\Alph{figure}}
\renewcommand{\thetable}{\Alph{table}}
\section{Additional matching results}
\label{suppsec:moreNIOM}
\begin{figure}[tb]
  \centering
  \includegraphics[width=\linewidth]{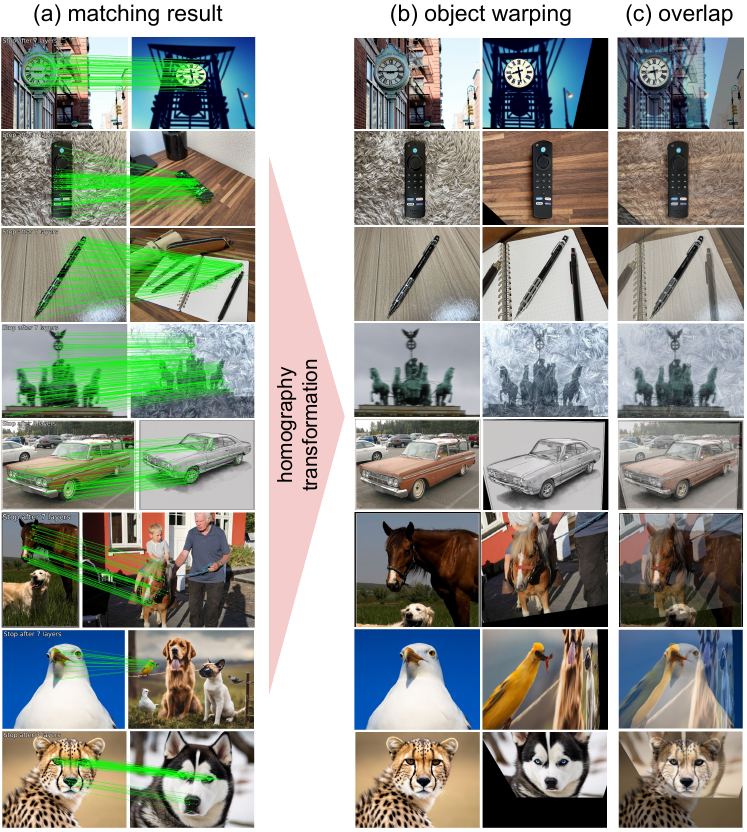}
  \caption{Object warping by homography transformation for fine-grained and consistent matching using our method. (b) By warping the object in the right image to the object position in the left image and overlaying the two images, (c) it can be confirmed that the object positions are superimposed, showing consistent alignment.
  }
  \label{fig:sup_homography}
\end{figure}

In~\cref{sub:exp_NIOM} of the main text, we evaluated non-identical object matching by inserting our method between SuperPoint~\cite{SuperPoint} and LightGlue~\cite{LightGlue}.
Here, ``SEW'' in the results refers to the case where only our weighting module is used without applying the Non-visual Object Pairing.
The matching results for various image pairs show that our method enhances matchers and provides dense and consistent correspondences.
The fine-grained correspondence can be applied to tasks such as object warping, as illustrated in~\cref{fig:sup_homography}.
To demonstrate this generality for various image pairs, we show more results in three cases: (i) same class (i.e., $y_A = y_B, D_A = D_B$), (ii) class discrepancy (i.e., $y_A \ne y_B, D_A = D_B$), (iii) domain shift (i.e., $o_A \equiv o_B$, $D_A\ne D_B$).
Additionally, we present more results for complex scenes containing multiple objects in a single image.
\paragraph{Setup.}
We collected various images with objects from the COCO~\cite{cocodataset} and ImageNet~\cite{imagenet} datasets. To prepare drawing or sketch versions of several images, we used the DreamStudio\footnote{DreamStudio: \url{https://beta.dreamstudio.ai/generate}}, which is backended by the Stable Diffusion~\cite{stable_diffusion}.
For details of the experiment, refer to~\cref{sub:exp_NIOM}. 
\subsection{Success cases}
\begin{figure*}[tb]
    \centering
    \includegraphics[width=0.8\linewidth]{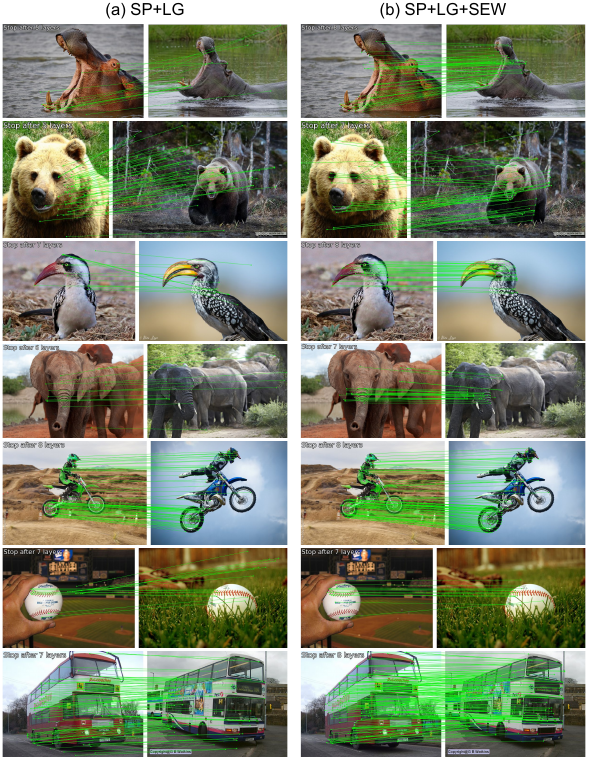}
    \caption{Matching results in the same class case.
    Image pairs show objects of the same class in ImageNet~\cite{imagenet} or COCO~\cite{cocodataset} datasets.
    (a) A combination of SuperPoint (SP)~\cite{SuperPoint} and LightGlue (LG)~\cite{LightGlue} only finds a small number of correspondences, and many of them are incorrect. (b) The proposed method SEW significantly improves the matching and performs dense and consistent correspondences.
    }
    \label{fig:sup_same}
\end{figure*}

\begin{figure*}[tb]
    \centering
    \includegraphics[width=0.8\linewidth]{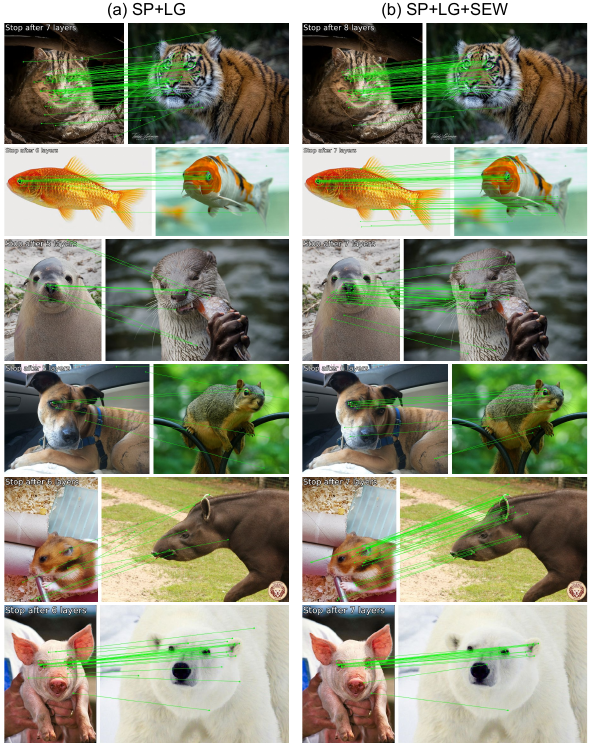}
    \caption{Matching results in the class discrepancy case.
    Image pairs show objects with class discrepancies in ImageNet~\cite{imagenet} or COCO~\cite{cocodataset} datasets.
    (a) A combination of SuperPoint (SP)~\cite{SuperPoint} and LightGlue (LG)~\cite{LightGlue} finds many inconsistent and scattered matches, and it is unclear which is the correct correspondence. (b) The proposed method SEW finds almost all matches for the corresponding parts of the animal, which largely improves the matching to consistency.
    }
    \label{fig:sup_discre}
\end{figure*}

\begin{figure*}[tb]
    \centering
    \includegraphics[width=0.8\linewidth]{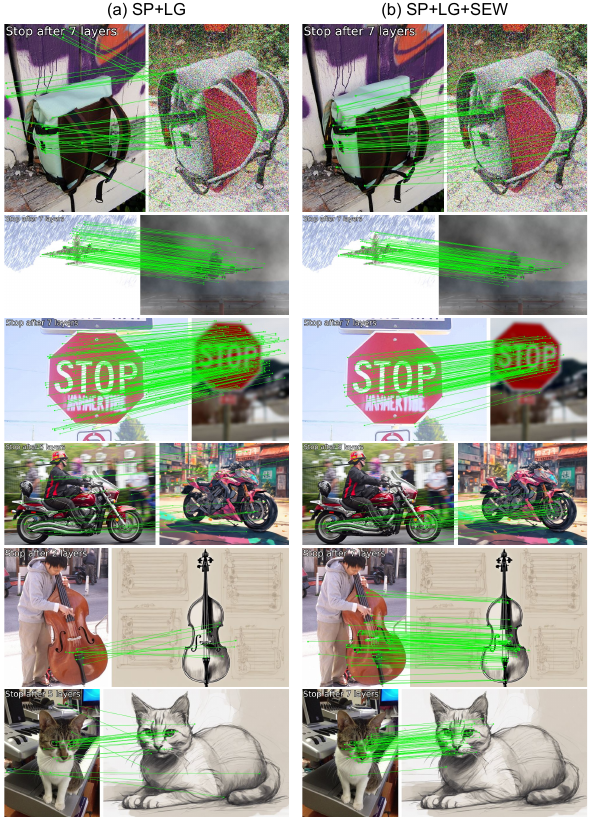}
    \caption{Matching results in the domain shift case.
    Image pairs for matching are ImageNet~\cite{imagenet} or COCO~\cite{cocodataset} datasets with common corruptions~\cite{CommonCorruption}, and illustrations or sketches of a motorcycle, a violin, and a cat are generated by the Stable Diffusion~\cite{stable_diffusion}.
    (a) A combination of SuperPoint (SP)~\cite{SuperPoint} and LightGlue (LG)~\cite{LightGlue} finds correct matching but also finds mismatches with many non-correspondence areas.
    (b) The proposed method SEW reduces these mismatches and further improves to denser matching between objects.
    }
    \label{fig:sup_domain}
\end{figure*}

\begin{figure*}[tb]
    \centering
    \includegraphics[width=0.8\linewidth]{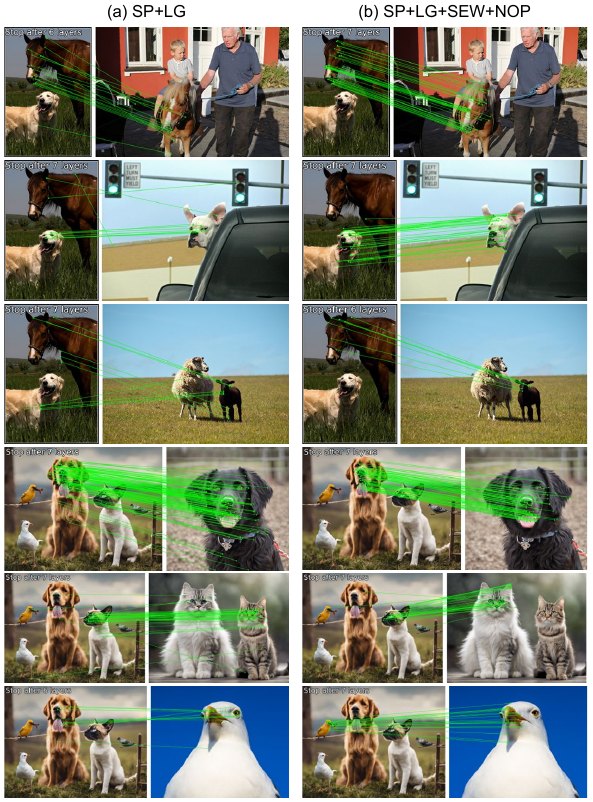}
    \caption{Matching results in complex scenes containing multiple objects.
    (a) A combination of SuperPoint (SP)~\cite{SuperPoint} and LightGlue (LG)~\cite{LightGlue} causes correspondences to scatter across multiple objects. (b) The proposed Non-visual Object Pairing (NOP) effectively selects semantically similar object pairs and provides condensed and consistent matching between the relevant parts of the objects.
    }
    \label{fig:sup_multiple}
\end{figure*}

Figure~\ref{fig:sup_same} shows the matching results for the same class case. As in the main text, The introduction of the proposed method turns sparse and inconsistent matching into dense and consistent one.
The combination of SuperPoint and LightGlue with our method is robust to object color and shape differences and performs dense and consistent matching.
Even in the case with class discrepancy, as shown in~\cref{fig:sup_discre}, the proposed method successfully corresponds the same parts (e.g., eyes to eyes, nose to nose) between various animals.
Figure~\ref{fig:sup_domain} shows the case of domain shift between images, where our method reduces mismatching caused by image corruptions and image style changes (e.g., photos and drawings).
Figure~\ref{fig:sup_multiple} shows cases where multiple objects are present in an image, demonstrating that our Non-visual Object Pairing selects semantically similar object pairs and provides condensed matching between the two objects.
These results indicate that the proposed method can perform accurate matching in many cases under the challenging condition of non-identical object matching.
\subsection{Failure cases}
\begin{figure*}[tb]
    \centering
    \includegraphics[width=0.8\linewidth]{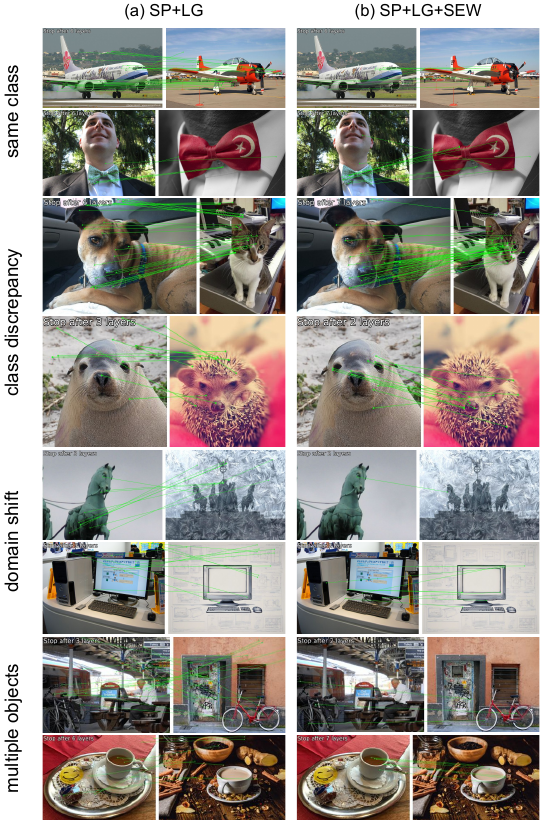}
    \caption{Failure cases of non-identical object matching.
    (a) A combination of SuperPoint (SP)~\cite{SuperPoint} and LightGlue (LG)~\cite{LightGlue} fails in matching.
    (b) The proposed method SEW improves several matching but still leaves many corresponding areas unmatched.
    }
    \label{fig:sup_fail}
\end{figure*}

There are some cases of failure in non-identical object matching, as shown in~\cref{fig:sup_fail}.
It is difficult to match objects that are of the same class but have significantly different shapes (e.g., the presence or absence of an airplane propeller) or objects that are similar in color or size to other parts (e.g., eyes and nose that are small, round, and black) of the object.
Furthermore, image corruptions sometimes hide the texture of the objects, making it difficult to pinpoint the areas to be matched.

\setlength{\tabcolsep}{8pt}
\begin{table*}[t]
    \centering
    \footnotesize
    \caption{Ablation study on the presence or absence of Non-visual Object Pairing in our method (Weighting Module: SEW, Non-visual Object Pairing: NOP).
    The pose accuracy (AUC) at the maximum angular error of $20\tcdegree$ of the relative pose estimation from image pairs MegaDepth-1500~\cite{LoFTR} under common corruptions~\cite{CommonCorruption}. 
    The left side shows both images corrupted with the same type of corruption, while the right side shows only one of the input images corrupted and the other as a clean image.
    NOP selects similar objects between images, but the MegaDepth-1500 dataset assumes that the same object appears in both images. Therefore, there is little need to select the object to be matched, resulting in minimal accuracy differences with and without NOP.
    }
    \label{tab:CleanCommonAbrationCSA}
    \begin{tabular}{lccc||ccc}
        \toprule
        \multirow{3.5}{*}{Common Corruptions} & 
        \multicolumn{3}{c}{AUC@$20\tcdegree$ (\textbf{corrupted} images)} & 
        \multicolumn{3}{c}{AUC@$20\tcdegree$ (\textbf{clean and corrupted} images)} \\ \cmidrule{2-4} \cmidrule{5-7}
        & 
        \multicolumn{3}{c}{\textit{keypoint detector : SuperPoint}} &
        \multicolumn{3}{c}{\textit{keypoint detector : SuperPoint}} \\
        &
        \multicolumn{1}{c}{LG} &
        \multicolumn{1}{c}{LG+SEW} &
        \multicolumn{1}{c}{LG+SEW+NOP} &
        \multicolumn{1}{c}{LG} &
        \multicolumn{1}{c}{LG+SEW} &
        \multicolumn{1}{c}{LG+SEW+NOP} \\ \midrule
        None (Clean)         & \textbf{80.61}  & 78.42 & 78.91 & \textbf{80.61}  & 78.42 & 78.91 \\ 
        \midrule
        Gaussian Noise       & 43.09  & \textbf{53.27}  & 52.21  & 27.54  & \textbf{41.01}  & 40.00  \\
        Shot Noise           & 43.41  & \textbf{53.82}  & 52.81  & 32.10  & \textbf{42.35}  & 41.41  \\
        Impulse Noise        & 44.98  & \textbf{50.11}  & 48.94  & 35.95  & \textbf{43.67}  & 42.65  \\
        Defocus Blur         & 32.69  & \textbf{48.35}  & 47.37  & 18.25  & \textbf{32.13}  & 31.25  \\
        Frosted Glass Blur   & 33.37  & \textbf{47.88}  & 47.25  & 34.25  & \textbf{44.80}  & 44.29  \\
        Motion Blur          & 42.12  & \textbf{53.63}  & 52.95  & 50.07  & \textbf{53.10}  & 52.46  \\
        Zoom Blur            & 24.40  & \textbf{31.21}  & 30.39  & 34.67  & \textbf{40.04}  & 39.36  \\
        Snow                 & \textbf{31.51}  & 30.33  & 30.40  & 56.24  & \textbf{58.15}  & 58.00  \\
        Frost                & \textbf{32.24}  & 31.71  & 31.83  & 62.20  & \textbf{64.87}  & 64.74  \\
        Fog                  & 70.99  & \textbf{73.49}  & 72.41  & 76.96  & \textbf{78.40}  & 77.37  \\
        Brightness           & \textbf{75.48}  & 75.08  & 75.21  & \textbf{77.14}  & 76.98  & 77.06  \\
        Contrast             & \textbf{39.50}  & 38.47  & 38.77  & 43.09  & \textbf{45.17}  & 45.11  \\
        Elastic Transform    & 54.78  & \textbf{66.21}  & 65.69  & 64.93  & \textbf{68.34}  & 67.92  \\
        Pixelate             & 67.94  & \textbf{68.24}  & 68.12  & 66.80  & \textbf{68.43}  & 68.37  \\
        JPEG Compression     & 29.54  & \textbf{36.97}  & 36.47  & 47.41  & \textbf{53.82}  & 53.45  \\  \midrule
        Average              & 44.40  & \textbf{50.58}  & 50.05  & 48.51  & \textbf{54.08}  & 53.56  \\ \bottomrule
    \end{tabular}
\end{table*}
\section{Robust image matching details}
With the same setup in~\cref{sub:exp_Pose} of the main text, we evaluate the effectiveness of the proposed method for LightGlue~\cite{LightGlue} and GlueStick~\cite{GlueStick} on the robustness to image corruptions and domain shift in image matching. 
First, we present an ablation study on the presence or absence of the Non-visual Object Pairing.
Second, we consider the maximum angular error at $10\tcdegree$ and $5\tcdegree$ instead of $20\tcdegree$.
The results of relative pose estimation with common corruptions~\cite{CommonCorruption} on one or both of the input images in the MegaDepth-1500~\cite{LoFTR} show that the proposed method improves the robustness of LightGlue and GlueStick.
In all experiments, we used the pretrained sparse matchers LightGlue~\cite{LightGlue}, and GlueStick~\cite{GlueStick}, and the dense matchers LoFTR~\cite{LoFTR} and EfficientLoFTR~\cite{EfficientLoFTR}, downloaded from their official repositories.\footnote{
LightGlue: \url{https://github.com/cvg/LightGlue}\\
GlueStick: \url{https://github.com/cvg/GlueStick}\\
LoFTR: \url{https://github.com/zju3dv/LoFTR}\\
ELoFTR: \url{https://github.com/zju3dv/efficientloftr}}  
\subsection{Abration on Non-visual Object Pairing} \label{subsec:csa}
We conducted an ablation study to assess the impact of our Non-visual Object Pairing (NOP). 
Recall that the NOP assigns object pairs based on the highest cosine similarity of 
text embeddings (CLIP) derived from their class labels, thereby focusing the Grad-CAM-based weighting on semantically matched objects. 
Table~\ref{subsec:csa} presents the robustness evaluation against common corruptions applied either to both images or only one of the input images, comPairing results with and without the NOP. We find that the accuracy does not vary significantly between these settings. This implies that, on the MegaDepth-1500 dataset, where image pairs typically depict the same physical object—our Grad-CAM-based weighting already assigns appropriate focus without needing explicit object selection by the NOP. Hence, the NOP's influence is less pronounced on standard image matching datasets with identical objects. In contrast, its benefits become more evident for non-identical object matching tasks, where semantically similar but distinct objects require a more careful pairing mechanism.
\subsection{Robustness against common corruptions} \label{subsec:detailsCC}
In this experiment, common corruptions are added to both input images to demonstrate robustness against corruption.
Table~\ref{tab:CC10and5} shows the results of relative pose estimation for maximum angular error of $10\tcdegree$ and $5\tcdegree$.
As the results show, the proposed method makes the LightGlue and GlueStick robust against most types of common corruptions~(roughly, 5\% to 10\% increase) with a slight decrease in clean accuracy. 
For the Noise category, LightGlue and GlueStick with our method even surpass dense matchers in both cases.
\subsection{Robustness against environmental changes} \label{subsec:detailsCleanCommon}
In contrast to~\cref{subsec:detailsCC}, common corruptions are added to one of the image pairs to demonstrate robustness against domain shift between images.
Table~\ref{tab:CleanCommon10and5} shows the results of relative pose estimation for maximum angular error of $10\tcdegree$ and $5\tcdegree$.
As the results show, the proposed method is robust to LightGlue and GlueStick, and in particular, significantly improves the AUC of the former for all Noise, Blur, and Digital categories, and the latter for all Noise and Blur categories.
For the Noise category, LightGlue and GlueStick with our method even exceed dense matchers in both cases.

\section{Other semantic models for weighting}
\label{suppsec:other_semantic_models}

In our main experiments, we adopted YOLOv7~\cite{YOLOv7} and Grad-CAM~\cite{GradCAM} as a fast and practical choice for generating semantic heatmaps. Here, we examine alternative semantic models and discuss their effectiveness as weighting sources for descriptors.

\begin{figure}[tb]
  \centering
  \includegraphics[width=1.0\linewidth]{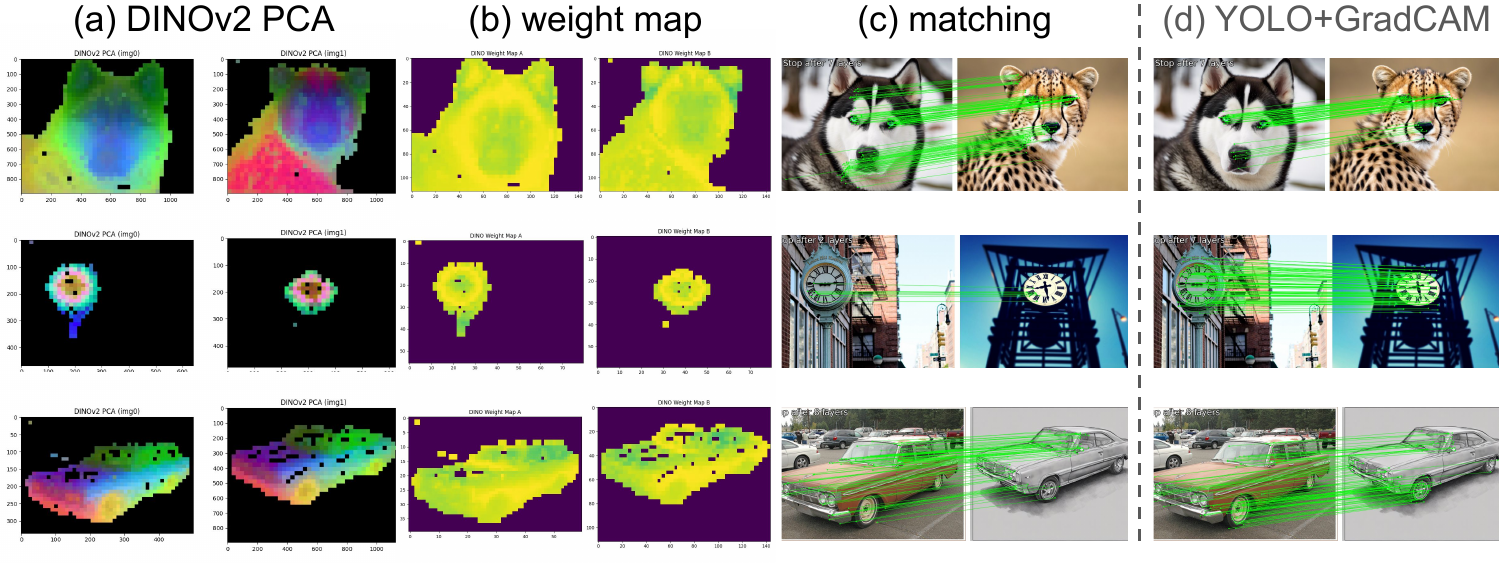}
  \caption{Comparison of weighting. (a) Semantic visualization using PCA of DINOv2 features. (b) Corresponding weight maps used to reweight descriptors. (c) Matching results with DINOv2-based weights. (d) Matching results with YOLOv7 and Grad-CAM (baseline). The DINOv2-based weights do not necessarily improve the performance compared to the original configuration.}
  \label{fig:sup_dino}
\end{figure}

\paragraph{DINOv2 as an alternative.}
We tested DINOv2~\cite{dinov2}, a vision foundation model, to generate semantic representations. First, we computed principal components of the DINOv2 features for each image to visualize the semantic regions. Figure~\ref{fig:sup_dino} (a) shows the PCA-based visualization, which captures semantic regions to some extent. Then, we generated weight maps by normalizing the PCA scores, as shown in Fig.~\ref{fig:sup_dino} (b). These weight maps were used to reweight keypoint descriptors, following the same procedure as our Grad-CAM-based weighting.

\paragraph{Effect on matching.}
Figure~\ref{fig:sup_dino} (c) illustrates the matching results when the DINOv2-based weight maps were applied. Compared to the results using YOLOv7 and Grad-CAM (Fig.~\ref{fig:sup_dino} (d)), the DINOv2-based weighting did not consistently improve the matching performance. In some cases, it even degraded the quality due to overly smooth attention over the image or misaligned semantic focus.
These results indicate that strong semantic models do not always translate to better weighting for sparse feature matching. While DINOv2 performs well in recognition or segmentation tasks, its internal features may not align well with fine-grained keypoint matching needs. 

\setlength{\tabcolsep}{3pt}
\begin{table}[t]
    \centering
    \footnotesize
    \caption{
    Comparison with a semantic correspondence method. We report AUC@20$^\circ$ of relative pose estimation using GeoAware-SC~\cite{GeoAwareSC} and our method (SuperPoint + LightGlue + SEW) on image pairs under common corruptions. The left columns show results when both images are corrupted, while the right columns show results when only one image is corrupted (domain shift setup). Although both methods perform similarly on clean images, our method consistently outperforms GeoAware-SC under corrupted settings while being more than 10 times faster.}
    \label{tab:geoaware_ours}
    \begin{tabular}{lcc||cc}
        \toprule
        \multirow{2.5}{*}{Common Corruptions} &
        \multicolumn{2}{c||}{\textbf{corrupted} pairs} &
        \multicolumn{2}{c}{\textbf{clean / corrupted} pairs} \\ \cmidrule{2-3} \cmidrule{4-5}
        & GeoAware-SC & Ours & GeoAware-SC & Ours \\ \midrule
        None (clean)        & 55.28 & \textbf{78.42} & 55.28 & \textbf{78.42} \\
        \midrule
        gaussian\_noise     & 29.17 & \textbf{53.27} & 33.57 & \textbf{41.01} \\
        shot\_noise         & 33.82 & \textbf{53.82} & 36.49 & \textbf{42.35} \\
        impulse\_noise      & 31.39 & \textbf{50.11} & 34.82 & \textbf{43.67} \\
        defocus\_blur       & 42.32 & \textbf{48.35} & 26.51 & \textbf{32.13} \\
        glass\_blur         & 39.15 & \textbf{47.88} & 39.83 & \textbf{44.80} \\
        motion\_blur        & 39.97 & \textbf{53.63} & 39.36 & \textbf{53.10} \\
        zoom\_blur          & 11.37 & \textbf{21.31} & 20.78 & \textbf{40.04} \\
        snow                & 18.44 & \textbf{30.33} & 35.60 & \textbf{58.15} \\
        frost               & 10.47 & \textbf{31.71} & 32.52 & \textbf{64.87} \\
        fog                 & 48.02 & \textbf{73.49} & 51.31 & \textbf{78.40} \\
        brightness          & 49.40 & \textbf{75.08} & 53.48 & \textbf{76.98} \\
        contrast            & 34.55 & \textbf{38.47} & 29.58 & \textbf{45.17} \\
        elastic\_transform  & 40.33 & \textbf{66.21} & 42.75 & \textbf{68.34} \\
        pixelate            & 48.20 & \textbf{68.24} & 48.92 & \textbf{68.43} \\
        jpeg\_compression   & 33.06 & \textbf{36.97} & 42.61 & \textbf{53.82} \\
        \midrule
        Average             & 33.98 & \textbf{50.82} & 37.88 & \textbf{54.08} \\
        Time [ms/pair]      & 710.36 & \textbf{68.43} & 715.24 & \textbf{68.37} \\
        \bottomrule
    \end{tabular}
\end{table}
\section{Matching algorithm with Geo-Aware-SC} \label{suppsec:GeoAwareMatchingAlgorithm}
In this section, we explain how semantic correspondence is leveraged within the Geo-Aware-SC~\cite{GeoAwareSC} framework to achieve feature matching. We introduce and compare two approaches: a Random Approach and a Heatmap-Based Approach. The first method samples pixel locations uniformly, while the second method uses a precomputed heatmap to prioritize semantically meaningful regions.

\subsection{Random sampling approach}
This approach uniformly selects a fixed number of points from the valid image region. 
In this random approach, each image is first resized to a fixed square resolution while preserving its aspect ratio. Any remaining space is padded to maintain the square shape and a binary mask is generated to indicate the valid (non-padded) region.
Next, we apply bilinear interpolation to upsample the feature maps so that each pixel in the resized image can be associated with a distinct feature vector. 
From the valid region, we randomly select a fixed number of pixels for the source image and extract their feature descriptors, which are then normalized.
The target image undergoes the same resizing and upsampling procedure, but instead of random selection, we use the entire valid region. 
We compute the cosine similarity between every descriptor of the source image and all descriptors of the target image. For each descriptor in the source image, the location in the target image with the highest similarity is identified as the match. 
Finally, these correspondences are visualized side-by-side.

\subsection{Heatmap-based sampling approach}
This approach leverages a precomputed semantic heatmap to guide the sampling process, ensuring that pixels in semantically important regions are more likely to be selected.
The heatmap-based method follows a similar resizing and upsampling procedure, but it incorporates a precomputed heatmap to guide the selection of sample points in the source image. 
The heatmap is aligned to the same resolution as the resized image and normalized. Each valid pixel in the source image is associated with a heatmap value, which serves as a sampling probability. 
This probabilistic selection ensures that pixels in regions with higher semantic importance are more likely to be chosen.
After selecting the sample points based on the heatmap, the descriptors are extracted and normalized just as in the random method. 
The target image also has its feature maps upsampled, and the descriptors are computed for its valid region. Each descriptor in the source image is then matched to the most similar descriptor in the target image, allowing important regions indicated by the heatmap to be prioritized during matching.
\subsection{Comparison with Semantic Matching Methods} \label{subsec:geoaware_comparison}
To further contextualize the effectiveness of our method, we compared it with GeoAware-SC~\cite{GeoAwareSC}, a representative semantic correspondence method. Although semantic matching techniques such as GeoAware-SC are typically designed for high-level alignment of semantically similar regions rather than fine-grained keypoint-level matching, they serve as a useful reference for evaluating robustness under challenging conditions.

The experimental setup followed that of~\cref{sub:exp_Pose}, focusing on relative pose estimation under common corruptions. As presented in~\cref{tab:geoaware_ours}, our method achieves significantly higher AUC@20$^\circ$ scores for both corrupted image pairs (50.82\% vs.\ 33.98\%) and clean-corrupted image pairs (54.08\% vs.\ 37.88\%) compared to GeoAware-SC. Moreover, the average inference time per image pair is an order of magnitude faster (68~ms vs.\ 710~ms), indicating practical advantages in efficiency.

\begin{figure*}[tb]
      \centering
      \includegraphics[width=0.8\linewidth]{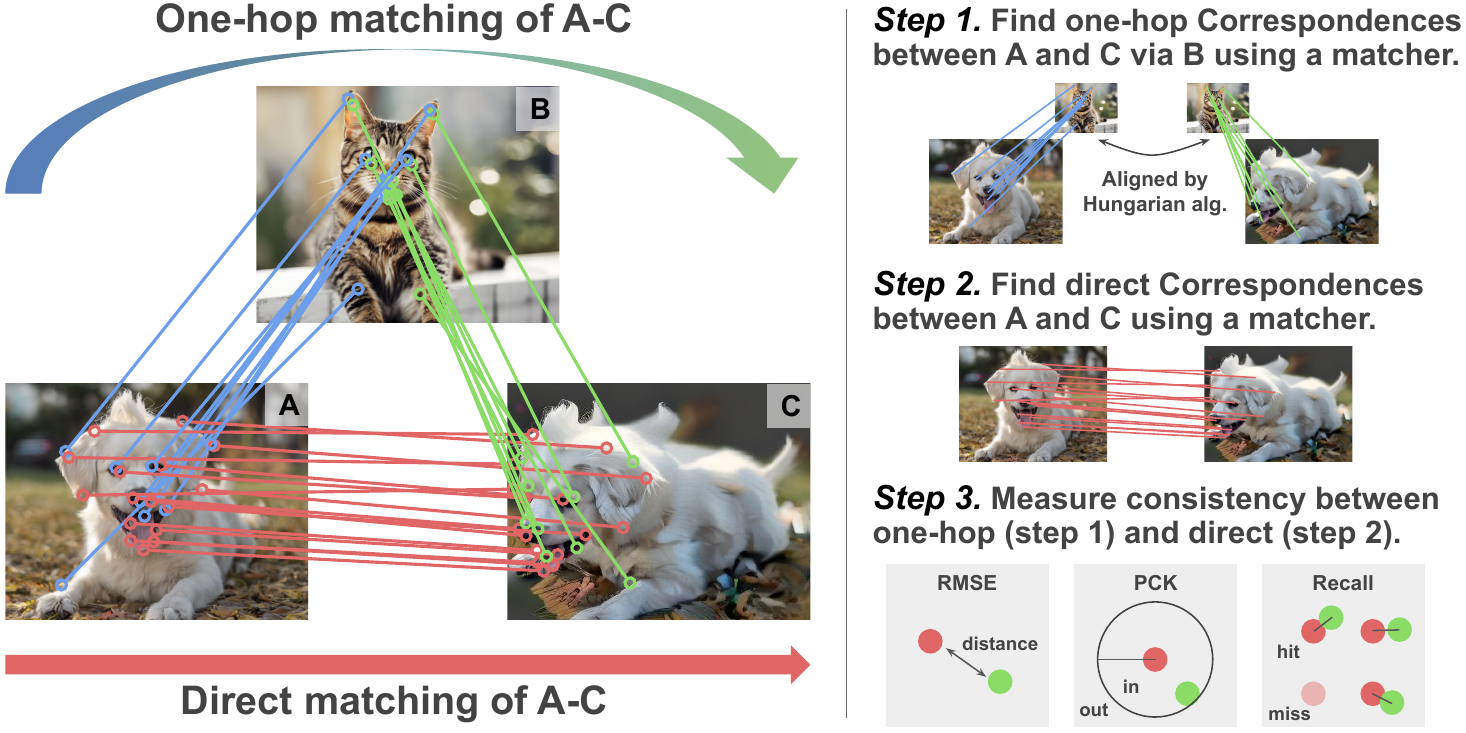}
      \caption{\textbf{Triangular Matching Consistency (TMC) overview.} Given two images $A$ and $C$ of the same object with different viewpoints (where reliable direct correspondences can be defined), and another semantically similar but non-identical object $B$, we compare the direct correspondences between $(A,C)$ with the one-hop correspondences $A{\leftrightarrow}B{\leftrightarrow}C$. A method with high TMC produces one-hop matches that closely agree with the direct baseline, thereby enabling annotation-free quantitative evaluation of non-identical object matching.}
      \label{fig:TMC_overview}
\end{figure*}

\section{Triangular Matching Consistency}
\label{supp_TMC_details}
We propose \emph{Triangular Matching Consistency} (TMC), a new annotation-free metric for quantitatively evaluating non-identical object matching. 
Defining ground-truth correspondences between non-identical objects is inherently difficult, as fine-grained pixel-level matches are often ambiguous. 
TMC avoids this limitation by measuring the consistency between the following two matchings: Letting $A,B,C$ be images of non-identical objects, we compute
\begin{itemize}[labelindent=10pt]
    \item direct matching of $(A,C)$
    \item one-hop matching through $(A,B)$ and $(B,C)$,
\end{itemize}
Images $A,C$ are expected to be very similar, and thus their direct correspondences can serve as a reliable reference. A non-identical object matching method is expected to approximate them even via the one-hop matching via semantically similar $B$.
This design enables quantitative evaluation without annotations and naturally captures cross-object coherence.
An overview of this procedure is shown in Fig.~\ref{fig:TMC_overview}.
TMC reports three complementary aspects of performance: (i) error magnitude (RMSE), (ii) accuracy as the fraction of correct correspondences (PCK), and (iii) coverage of the baseline correspondences (Recall). 
We next describe the procedure step by step.

\subsection{Direct and one-hop matchings}
Let $\cA$ be the matching algorithm to evaluate. Given images $A, C$, this returns a matching, 
\begin{align}
    \cA(A, C) = \qty{(\bm{p}_i^A, \bm{p}_i^C)}_{i=1}^{M_1},
    \label{eq_tmc_directmatch}
\end{align}
where $\bm{p}_i^A, \bm{p}_i^C \in [0, 1]^2$ are the matched position vectors (i.e., normalized coordinates) in $A$ and $C$, respectively, and $M_1$ is the number of matched pairs. Note that each position vector (say, $\bm{p}_i^A)$ in the matching has an associated descriptor (say, $\bm{d}_i^A)$. 
We denote the one-hop matching via $B$ by 
\begin{align}
    \cA(A, C; B) = \qty{(\widehat{\bm{p}}_i^A, \widehat{\bm{p}}_i^C)}_{i=1}^{M_2},
\end{align}
where $M_2$ is the number of matched pairs, and this is obtained by combining 
\begin{align}
    \cA(A, B) &= \qty{(\bm{q}_i^A,  
    \bm{q}_i^B)}_{i=1}^{N_1}, \\
    \cA(B, C) &= \qty{(\bm{r}_i^B, \bm{r}_i^C)}_{i=1}^{N_2}.
\end{align}
Particularly, the Hungarian algorithm~\cite{hungarian} is applied to pair position vectors in ${\bm{q}_i^B}$ and ${\bm{r}_i^B}$, through which the one-hop matching is found. For example, if the Hungarian algorithm pairs $(\bm{q}_i^B, \bm{r}_j^B)$, then we have $(\bm{q}_i^A, \bm{r}_j^C) =: (\widehat{\bm{p}}_i^A, \widehat{\bm{p}}_i^C)$. 
The implementation details are as follows. 
\paragraph{Outlier removal.} 
The outlier removal by RANSAC-based geometric verification is common in the literature to obtain reliable inlier matches~\cite{RANSAC,MegaDepth, LoFTR, LightGlue}. In TCM, the removal step is inserted before obtaining \cref{eq_tmc_directmatch}. 
Particularly, given the initial matching between A and C, geometric verification is applied via RANSAC to estimate either a homography or a fundamental matrix, depending on the scene geometry, and only inliers are retained. 
However, the outlier removal is not applied to $\cA(A,B)$ and $\cA(B,C)$ because matching between non-identical objects can be regarded as ``outliers''; removing them could discard semantically meaningful pairs.

\paragraph{Proximity used in Hungarian algorithm.} 
The image B has two sets of position vectors, $\{\bm q_i^B\}_{i=1}^{N_1}$ from $\cA(A, B)$ and $\{\bm r_i^B\}_{i=1}^{N_2}$ from $\cA(B, C)$. The Hungarian algorithm determines the alignment of $M_2 := \min\{N_1, N_2\}$ vectors between them by minimizing the cost. The cost is defined as follows. First, the geometric and descriptor discrepancies are 
\begin{align}
    \delta^{p}_{ij} &= \|\bm q_i^B-\bm r_j^B\|, \\
    \delta^{d}_{ij} &= \|\bm{e}_i^B-\bm{f}_j^B\|,
\end{align}
where $\bm{e}_i^B, \bm{f}_j^B$ are the descriptors associated with $\bm{q}_i^B, \bm{r}_j^B$, respectively.
The cost is then defined by
\begin{align}
    c_{ij} &= \lambda_p \, \delta^{p}_{ij}
           + \lambda_d \, \delta^{d}_{ij},
\end{align}
where $\lambda_p,\lambda_d \ge 0$ are hyper-parameters.
In our experiments, we set $\lambda_p = 1.0$ and $\lambda_d = 0.3$.

\subsection{Measuring matching consistency }
Using the direct and one-hop matching, we now measure the matching consistency by three complementary metrics. 
The consistency is measured using the position vectors of the image $C$. Recall that we have $\qty{\bm{p}_i^C}_{i=1}^{M_1}$ from the direct matching and $\qty{\widehat{\bm{p}}_i^C}_{i=1}^{M_2}$ from the one-hop matching. 
Let 
\begin{align}
    \Delta_{ij} = \|\bm{p}^C_{i} - \widehat{\bm{p}}^C_{i}\|_2,\quad i = 1,\ldots, M_1,\ j=1,\ldots, M_2,
\end{align}
and we introduce three metrics of consistency as follows. 

\medskip\noindent\textbf{RMSE (Root Mean Squared Error).}
RMSE measures the magnitude of geometric error between the direct and one-hop matching. 
\begin{align}
    \mathrm{RMSE}
    &= \sqrt{\frac{1}{M_2}\sum_{j=1}^{M_2} \min_{i}\Delta_{ij}^2}.
\end{align}
Noting that the position vectors are normalized, its value ranges in $[0,\sqrt{2}]$. Smaller values indicate better consistency.

\medskip\noindent\textbf{PCK (Percentage of Correct Keypoints).}
PCK~\cite{PFPASCAL} measures the proportion of accurate correspondences within a tolerance $\tau>0$ in the normalized plane:
\begin{align}
    \mathrm{PCK}(\tau)
    &= \frac{1}{M_2}\sum_{j=1}^{M_2}\mathbf{1}[\min_{i}\Delta_{ij} \le \tau],
\end{align}
where $\mathbf{1}[\,\cdot\,]$ is the indicator function. High values at small $\tau$ indicate many highly accurate pairs.
In our experiments, $\tau$ is set to $0.01$, $0.05$, and $0.1$.
To reduce sensitivity to any single threshold.
PCK takes values in $[0,1]$, and larger values indicate better consistency.

\medskip\noindent\textbf{Recall.}
Recall measures the coverage of the direct matching $A{\leftrightarrow}C$ correspondences by the one-hop correspondences $A{\leftrightarrow}B{\leftrightarrow}C$. 
\begin{align}
    \mathrm{Recall}(\tau)
    &= \frac{1}{M_1}\sum_{i=1}^{M_1}\mathbf{1}[\min_{j}\Delta_{ij} \le \tau],
\end{align}
In our experiments, $\tau$ is set to $0.01$, $0.05$, and $0.1$. 
Recall takes values in $[0,1]$, and larger values indicate better coverage.

\section{Additional TMC results}
\label{secsupp:moreTMCresults}
This section provides additional TMC results that complement the wild-category evaluation presented in Sec.~\ref{sub:exp_TMC}. 
We report per-class performance for the dog and cat categories of the Animal Faces dataset~\cite{animalfaces}. 
The evaluation protocol follows the same triplet construction and metrics described in the main paper: error magnitude (RMSE), accuracy (PCK), and coverage (Recall) at distance thresholds of 0.01, 0.05, and 0.10 of the image size.

Table~\ref{tab:experiment-tmc-dog-cat} shows, as in the wild-category results, our weighting consistently improves all matchers across all metrics. 
Sparse matchers combined with our module achieve substantially lower geometric error and higher accuracy than their baselines, and they also outperform dense detector-free methods in most settings. 
These class-wise results further confirm that our method enhances the robustness and consistency of non-identical object matching.

\setlength{\tabcolsep}{2pt}
\begin{table}[t]
    \centering
    \footnotesize
    \caption{
        Triangular Matching Consistency (TMC) results for the dog and cat categories 
        in the Animal Faces dataset~\cite{animalfaces}. 
        Sparse matchers (LightGlue (LG), GlueStick (GS)) with SuperPoint, and dense matchers 
        (LoFTR, Efficient LoFTR) are shown for reference.
        Metrics include RMSE (lower is better), PCK, and Recall at thresholds of 
        0.01, 0.05, and 0.10 of the image size.
    }
    \label{tab:experiment-tmc-dog-cat}
    \begin{tabular}{l l l c c c c c c c}
        \toprule
        \multirow{2}{*}{\shortstack{Class}} &
        \multicolumn{2}{c}{\multirow{2}{*}{Matching Method}} &
        \multirow{2}{*}{RMSE$\downarrow$} &
        \multicolumn{3}{c}{PCK$\uparrow$} &
        \multicolumn{3}{c}{Recall$\uparrow$} \\
        & \multicolumn{2}{c}{} & & 0.01 & 0.05 & 0.10 & 0.01 & 0.05 & 0.10 \\ 
        \midrule

        \multirow{6}{*}{\centering Dog} &
        \multirow{4}{*}{\shortstack[l]{Sparse \\ (+SuperPoint)}} 
            & LG
                & \multicolumn{1}{c|}{0.20}
                & 48.5 & 60.2 & \multicolumn{1}{c|}{66.1}
                & 51.3 & 62.4 & \multicolumn{1}{c}{67.8} \\

        & & LG+SEW
                & \multicolumn{1}{c|}{\textbf{0.15}}
                & \textbf{69.4} & \textbf{77.1} & \multicolumn{1}{c|}{\textbf{82.0}}
                & \textbf{69.4} & \textbf{76.7} & \multicolumn{1}{c}{\textbf{81.9}} \\ 

        & & GS
                & \multicolumn{1}{c|}{0.22}
                & 45.0 & 55.8 & \multicolumn{1}{c|}{60.3}
                & 47.9 & 57.2 & \multicolumn{1}{c}{61.8} \\

        & & GS+SEW
                & \multicolumn{1}{c|}{0.17}
                & 61.8 & 71.3 & \multicolumn{1}{c|}{77.4}
                & 60.4 & 70.1 & \multicolumn{1}{c}{76.2} \\ 
        \cmidrule{2-10}

        & \multirow{2}{*}{\shortstack[l]{Dense \\ (no detector)}} 
            & LoFTR
                & \multicolumn{1}{c|}{0.26}
                & 50.4 & 59.3 & \multicolumn{1}{c|}{63.7}
                & 48.7 & 56.1 & \multicolumn{1}{c}{60.9} \\

        & & ELoFTR
                & \multicolumn{1}{c|}{0.20}
                & 59.6 & 68.8 & \multicolumn{1}{c|}{72.5}
                & 58.4 & 67.5 & \multicolumn{1}{c}{71.2} \\ 
        \midrule

        \multirow{6}{*}{\centering Cat} &
        \multirow{4}{*}{\shortstack[l]{Sparse \\ (+SuperPoint)}} 
            & LG
                & \multicolumn{1}{c|}{0.13}
                & 56.8 & 72.1 & \multicolumn{1}{c|}{78.4}
                & 59.7 & 74.5 & \multicolumn{1}{c}{80.1} \\

        & & LG+SEW
                & \multicolumn{1}{c|}{\textbf{0.08}}
                & \textbf{71.2} & \textbf{84.6} & \multicolumn{1}{c|}{\textbf{90.4}}
                & \textbf{71.2} & \textbf{84.4} & \multicolumn{1}{c}{\textbf{90.2}} \\ 

        & & GS
                & \multicolumn{1}{c|}{0.15}
                & 52.7 & 67.8 & \multicolumn{1}{c|}{74.0}
                & 54.9 & 70.3 & \multicolumn{1}{c}{75.8} \\

        & & GS+SEW
                & \multicolumn{1}{c|}{0.10}
                & 66.5 & 80.3 & \multicolumn{1}{c|}{86.9}
                & 65.1 & 79.1 & \multicolumn{1}{c}{86.3} \\ 
        \cmidrule{2-10}

        & \multirow{2}{*}{\shortstack[l]{Dense \\ (no detector)}} 
            & LoFTR
                & \multicolumn{1}{c|}{0.17}
                & 55.3 & 70.2 & \multicolumn{1}{c|}{76.1}
                & 53.8 & 69.0 & \multicolumn{1}{c}{74.8} \\

        & & ELoFTR
                & \multicolumn{1}{c|}{0.12}
                & 63.9 & 78.1 & \multicolumn{1}{c|}{83.4}
                & 62.7 & 77.0 & \multicolumn{1}{c}{82.6} \\
        \bottomrule
    \end{tabular}
\end{table}

\section{Limitations}
\label{secsupp:limitations}
\begin{figure}[tb]
      \centering
      \includegraphics[width=\linewidth]{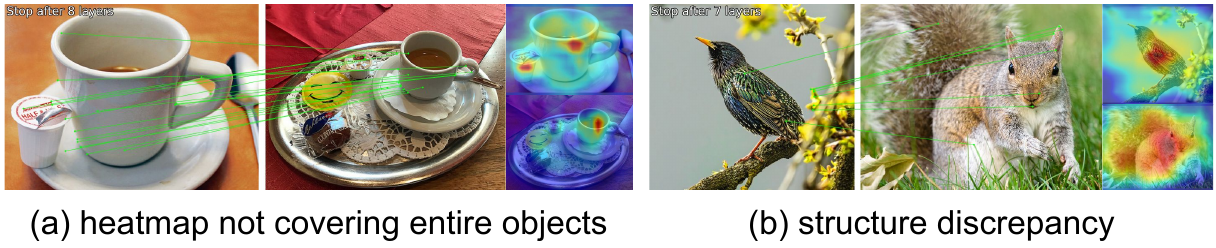}
      \caption{
      Failure cases: (a) high heatmap scores cover only a small region of objects, (b) the objects have clearly different structures.
      }
      \label{fig:exp_limit_bad}
\end{figure}

    Figure~\ref{fig:exp_limit_bad} shows failure cases caused by (a) heatmaps not covering a whole part of objects and (b) great structural differences between objects.
    In the former case, the importance scores take large only on particular parts of the objects. Further, the large importance scores on the irrelevant object disrupt the matching. This may be resolved using better visual explanation methods~\cite{learningDeep, GradCAMplus, AttentionRollout, SHAP, MoXI, VXCODE} or vision foundation model~\cite{dinov2}.
    For the latter case, objects with great structural differences cannot be matched as there is little correspondence between the objects, and eventually, regions of similar colors or textures are matched.
    To address these cases, one may introduce the vision and language model, such as the segment anything model~\cite{SegmentAnything, SegmentAnything2} to complement the visual information with text and part-level mask information.

\section{Related Work Details}
\label{secsupp:rel_details}
\medskip\noindent\textbf{Keypoint detectors.}
    Classically, keypoint detection was done using hand-crafted local features such as SIFT~\cite{SIFT} and ORB~\cite{ORB}. Recently, learning-based methods have been developed~\cite{SuperPoint, diffusionkeypoints, silk, featurebooster, XFeat, dedodev2}. 
    Particularly, SuperPoint~\cite{SuperPoint} and DISK~\cite{DISK} are two popular examples. The former is based on convolutional neural networks, and the latter is based on reinforcement learning. There is also a fully differentiable and lightweight method with a minimal model configuration~\cite{silk}.
    Unsupervised keypoint detection based on a text-to-image diffusion model~\cite{diffusionkeypoints} also exists. 

\medskip\noindent\textbf{Feature matchers.}
    Feature matchers~\cite{feature_matching_survey, feature_matching_survey2} find a fine-grained correspondence between images. 
    The feature matchers are categorized into two types based on the density of matching points: sparse matchers and dense matchers.
    The former runs faster, while the latter gives dense and high accuracy.
    The focus of our study lies in sparse matchers.  A seminal sparse matcher, SuperGlue~\cite{SuperGlue}, addresses the partial assignment problem for matching by integrating the Transformer attention mechanism~\cite{Transformer} with optimal transport~\cite{OptimalTransport}.
    The state-of-the-art sparse matcher, LightGlue~\cite{LightGlue}, improves SuperGlue by introducing a hierarchical matching structure, achieving high matching accuracy and real-time processing simultaneously. GlueStick~\cite{GlueStick} utilizes line segments for structural features.
    In contrast to sparse matchers, dense matchers~\cite{LoFTR, ASpanFormer, sparse_ncnet, patch2pix, cotr, roma, SGAM, ASTR, JamMa, SGAD} pursue matching accuracy rather than speed.
    Dense matchers input an image and infer matching using all pixels as keypoints.
    Therefore, it does not require a keypoint detector, so it is also called a detector-free model.
    LoFTR~\cite{LoFTR} proposed the first transformer-based method for dense matching, which achieves highly dense matching.
    Recent studies~\cite{ETO, MESA, RCM} such as Efficient LoFTR~\cite{EfficientLoFTR}, enabling fast and accurate matching by token aggregation, and ASTR~\cite{ASTR}, which focuses on the role of each pixel.

\medskip\noindent\textbf{Semantic correspondence models.}
    Semantic Correspondence~\cite{PFPASCAL, SCNet, SPair71k, cats, demystifyingSC, EffcientSemanticMatching, BringingViewpointGaps, SCacrossModalities} is a task of identifying a semantically similar point in the target image from the given point in the source image.
    Specifically, it enables the correspondence of parts, such as the mouth, nose, or right eye of dogs, between the source and target images. 
    The training of a model is conducted using a specialized dataset, such as SPair-71k~\cite{SPair71k} and PF-PASCAL~\cite{PFPASCAL}, which employ sparsely annotated part-level correspondences as supervisory data.
    Recently, GeoAware-SC~\cite{GeoAwareSC} has been proposed to integrate geometric and semantic information by leveraging pre-trained features from Stable Diffusion~\cite{stable_diffusion} and DINOv2~\cite{dinov2}, boosting semantic correspondence accuracy on major benchmarks~\cite{SPair71k, PFPASCAL}.
    Additionally, various models~\cite{GANSC, universalSC, SDDINO, DIFT, UFC, catsplus} have been proposed, including one that uses reverse diffusion feature maps~\cite{diffhyper} and unsupervised methods that learn semantic correspondence without annotated data~\cite{unsupervisedSC}. It is also worth noting that semantic correlation models run significantly slower than sparse matching methods because feature extraction depends on large generative models, such as  Stable Diffusion~\cite{stable_diffusion}. 
    For example, it runs roughly 10 times slower in Fig.~\ref{fig:task_NIOM}.

\setlength{\tabcolsep}{9pt}
\begin{table*}[b]
    \centering
    \caption{The pose accuracy (AUC) at the maximum angular error of $10\tcdegree$ and $5\tcdegree$ of the relative pose estimation from image pairs MegaDepth-1500~\cite{LoFTR} under common corruptions. Both images are corrupted with the same type of corruption. 
    Our method largely improves LightGlue (LG)~\cite{LightGlue} and GlueStick (GS)~\cite{GlueStick} for most categories and the average AUC. The results of a dense matchers, LoFTR~\cite{LoFTR} and Efficient LoFTR (ELoFTR)~\cite{EfficientLoFTR}, are shown for reference.
    Our method improves the robustness of sparse matchers with a marginal decrease in clean accuracy.
            }
    \label{tab:CC10and5}
    \begin{tabular}{lcc|cc|cc}
        \toprule
        \multirow{3.5}{*}{Common Corruptions} & \multicolumn{6}{c}{AUC@$10\tcdegree$ with pairs of \textbf{corrupted} images}     \\ \cmidrule{2-7}
        & 
        \multicolumn{4}{c}{\textit{keypoint detector : SuperPoint}} &
        \multicolumn{2}{c}{\textit{dense matcher}} \\
        &
        \multicolumn{1}{c}{LG} &
        \multicolumn{1}{c}{LG+SEW} &
        \multicolumn{1}{c}{GS} &
        \multicolumn{1}{c}{GS+SEW} &
        \multicolumn{1}{c}{LoFTR}  &
        \multicolumn{1}{c}{ELoFTR}  \\ \midrule
        None (Clean)         & \textbf{67.89}  & 66.98  & \textbf{64.11}  & 61.94  & 68.88 & 72.18  \\ \midrule
        Gaussian Noise       & 27.44  & \textbf{37.28}  & 28.64  & \textbf{35.11}  & 20.07 & 20.32  \\
        Shot Noise           & 27.07  & \textbf{39.79}  & 27.81  & \textbf{33.42}  & 24.27 & 23.89  \\
        Impulse Noise        & 28.70  & \textbf{37.25}  & 29.10  & \textbf{34.44}  & 22.27 & 23.75  \\
        Defocus Blur         & 18.58  & \textbf{24.51}  & 30.12  & \textbf{33.78}  & 43.27 & 40.33  \\
        Frosted Glass Blur   & 17.24  & \textbf{27.61}  & 29.16  & \textbf{31.75}  & 39.21 & 30.43  \\
        Motion Blur          & 27.08  & \textbf{34.85}  & 27.98  & \textbf{35.46}  & 40.68 & 38.01  \\
        Zoom Blur            & 12.50  & \textbf{20.23}  & 12.64  & \textbf{20.72}  & 12.01 & 13.66  \\
        Snow                 & \textbf{18.35}  & 12.56  & \textbf{12.46}  & 10.49  & 18.94 & 24.27  \\
        Frost                & \textbf{20.86}  & 18.75  & \textbf{17.64}  & 15.48  & 11.15 & 15.67  \\
        Fog                  & 56.32  & \textbf{62.62}  & 55.14  & \textbf{57.24}  & 53.05 & 59.85  \\
        Brightness           & 61.10  & \textbf{63.95}  & \textbf{58.75}  & 56.51  & 61.88 & 67.22  \\
        Contrast             & \textbf{24.33}  & 23.10  & 24.31  & \textbf{27.45}  & 38.57 & 44.78  \\
        Elastic Transform    & 36.76  & \textbf{47.02}  & 41.53  & \textbf{45.52}  & 41.27 & 42.79  \\
        Pixelate             & 51.60  & \textbf{60.66}  & \textbf{47.60}  & 47.32  & 62.55 & 64.85  \\
        JPEG Compression     & 16.60  & \textbf{27.73}  & 22.35  & \textbf{27.64}  & 43.49 & 43.50  \\ \midrule
        Average              & 29.64  & \textbf{35.86}  & 31.02  & \textbf{34.16}  & 35.51 & 36.89  \\ \midrule \midrule
        & \multicolumn{5}{c}{AUC@$5\tcdegree$ with pairs of \textbf{corrupted} images} \\ \midrule
        None (Clean)         & \textbf{50.43}  & 49.71  & \textbf{45.92}  & 43.11  & 52.52 & 56.38  \\ \midrule
        Gaussian Noise       & 14.70  & \textbf{24.44}  & 14.63  & \textbf{21.53}  & 9.85  & 9.57   \\ 
        Shot Noise           & 13.71  & \textbf{23.44}  & 13.44  & \textbf{18.06}  & 12.21 & 11.78  \\ 
        Impulse Noise        & 15.65  & \textbf{21.25}  & 16.90  & \textbf{20.64}  & 10.92 & 11.77  \\ 
        Defocus Blur         & 8.22   & \textbf{11.69}  & 16.59  & \textbf{20.64}  & 27.77 & 25.67  \\ 
        Frosted Glass Blur   & 7.17   & \textbf{18.09}  & 14.58  & \textbf{16.12}  & 24.59 & 16.17  \\ 
        Motion Blur          & 14.72  & \textbf{20.72}  & 15.26  & \textbf{22.34}  & 25.37 & 27.82  \\ 
        Zoom Blur            & 5.38   & \textbf{8.82}   & 6.02   & \textbf{14.16}  & 5.17  & 8.42   \\ 
        Snow                 & \textbf{9.48}   & 8.32   & \textbf{5.92}   & 3.87   & 9.91  & 13.79  \\ 
        Frost                & \textbf{11.92}  & 11.29  & \textbf{9.92}   & 7.72   & 6.09  & 8.75   \\ 
        Fog                  & 39.01  & \textbf{43.43}  & 35.88  & \textbf{37.58}  & 36.94 & 48.00  \\ 
        Brightness           & 43.60  & \textbf{47.98}  & \textbf{41.86}  & 39.60  & 45.13 & 50.73  \\ 
        Contrast             & 11.69  & \textbf{13.93}  & 12.75  & \textbf{15.29}  & 23.13 & 29.26  \\ 
        Elastic Transform    & 20.43  & \textbf{29.64}  & 24.29  & \textbf{27.28}  & 24.40 & 25.35  \\ 
        Pixelate             & 32.97  & \textbf{41.82}  & \textbf{29.86}  & 29.66  & 46.40 & 48.41  \\ 
        JPEG Compression     & 8.00   & \textbf{15.94}  & 9.71   & \textbf{14.79}  & 27.77 & 27.45  \\  \midrule
        Average              & 17.11  & \textbf{22.72}  & 17.84  & \textbf{20.62}  & 22.38 & 24.20  \\  \bottomrule
    \end{tabular}
\end{table*}
\setlength{\tabcolsep}{9pt}
\begin{table*}[b]
    \centering
    \caption{The pose accuracy (AUC) at the maximum angular error of $10\tcdegree$ and $5\tcdegree$ of the relative pose estimation from image pairs MegaDepth-1500~\cite{LoFTR} under common corruptions. 
    Only one of the input images is corrupted and the other is a clean image.
    Our method largely improves LightGlue (LG)~\cite{LightGlue} and GlueStick (GS)~\cite{GlueStick} for most categories and the average AUC. The results of dense matchers, LoFTR~\cite{LoFTR}, and Efficient LoFTR (ELoFTR)~\cite{EfficientLoFTR} are shown for reference.
    Our method improves the robustness of sparse matchers with a marginal decrease in clean accuracy.}
    \label{tab:CleanCommon10and5}
    \begin{tabular}{lcc|cc|cc}
        \toprule
        \multirow{3.5}{*}{Common Corruptions} & \multicolumn{6}{c}{AUC@$10\tcdegree$ with pairs of \textbf{clean and corrupted} images}     \\ \cmidrule{2-7}
        & 
        \multicolumn{4}{c}{\textit{keypoint detector : SuperPoint}} &
        \multicolumn{2}{c}{\textit{dense matcher}} \\
        &
        \multicolumn{1}{c}{LG} &
        \multicolumn{1}{c}{LG+SEW} &
        \multicolumn{1}{c}{GS} &
        \multicolumn{1}{c}{GS+SEW} &
        \multicolumn{1}{c}{LoFTR} &
        \multicolumn{1}{c}{ELoFTR}  \\ \midrule
        None (Clean)         & \textbf{67.89}  & 66.98  & \textbf{64.11}  & 61.94  & 68.88 & 72.18  \\ \midrule
        Gaussian Noise       & 16.63  & \textbf{28.34}  & 28.57  & \textbf{32.42}  & 24.17 & 27.01  \\
        Shot Noise           & 19.97  & \textbf{32.50}  & 26.23  & \textbf{29.92}  & 26.58 & 29.11  \\
        Impulse Noise        & 23.56  & \textbf{34.74}  & 33.12  & \textbf{36.26}  & 25.67 & 28.67  \\
        Defocus Blur         & 8.29   & \textbf{22.43}  & 12.03  & \textbf{18.17}  & 36.85 & 31.36  \\
        Frosted Glass Blur   & 19.00  & \textbf{32.78}  & 26.39  & \textbf{29.66}  & 40.40 & 34.59  \\
        Motion Blur          & 33.29  & \textbf{39.69}  & 27.88  & \textbf{38.30}  & 39.41 & 37.15  \\
        Zoom Blur            & 20.38  & \textbf{26.23}  & 16.76  & \textbf{25.67}  & 21.60 & 21.63  \\
        Snow                 & 39.74  & \textbf{41.07}  & 30.74  & \textbf{33.00}  & 35.91 & 43.33  \\
        Frost                & 47.61  & \textbf{51.78}  & 38.80  & \textbf{41.62}  & 32.97 & 38.00  \\
        Fog                  & 62.39  & \textbf{64.08}  & \textbf{61.11}  & 60.80  & 60.74 & 65.26  \\
        Brightness           & \textbf{63.40}  & 61.75  & \textbf{62.69}  & 60.04  & 65.64 & 69.24  \\
        Contrast             & 27.81  & \textbf{33.12}  & 25.86  & \textbf{28.95}  & 30.27 & 40.35  \\
        Elastic Transform    & 47.62  & \textbf{56.66}  & 52.17  & \textbf{55.08}  & 53.61 & 56.20  \\
        Pixelate             & 50.14  & \textbf{56.68}  & \textbf{50.33}  & 48.63  & 63.14 & 66.22  \\
        JPEG Compression     & 29.92  & \textbf{40.31}  & \textbf{40.93}  & 40.61  & 53.58 & 57.23  \\  \midrule
        Average              & 33.98  & \textbf{41.48}  & 35.57  & \textbf{38.61}  & 40.70 & 43.02  \\  \midrule \midrule
        & \multicolumn{5}{c}{AUC@$5\tcdegree$ with pairs of \textbf{clean and corrupted} images}     \\ \midrule
        None (Clean)         & \textbf{50.43}  & 49.71  & \textbf{45.92}  & 43.11  & 52.52 & 56.38  \\ \midrule
        Gaussian Noise       & 8.42   & \textbf{20.10}  & 16.51  & \textbf{20.52}  & 16.68 & 18.54  \\
        Shot Noise           & 10.30  & \textbf{20.52}  & 16.14  & \textbf{19.31}  & 18.14 & 19.72  \\
        Impulse Noise        & 12.88  & \textbf{24.11}  & 21.41  & \textbf{25.35}  & 17.75 & 19.55  \\
        Defocus Blur         & 3.22   & \textbf{17.92}  & 5.95   & \textbf{15.89}  & 24.73 & 21.55  \\
        Frosted Glass Blur   & 8.56   & \textbf{22.42}  & 15.54  & \textbf{18.79}  & 26.87 & 22.61  \\
        Motion Blur          & 18.95  & \textbf{26.22}  & 13.94  & \textbf{24.36}  & 25.92 & 28.77  \\
        Zoom Blur            & 9.81   & \textbf{19.27}  & 8.95   & \textbf{17.68}  & 15.03 & 17.16  \\
        Snow                 & 22.94  & \textbf{26.90}  & 17.27  & \textbf{20.56}  & 23.94 & 31.62  \\
        Frost                & 31.58  & \textbf{37.80}  & 24.13  & \textbf{27.95}  & 23.07 & 29.96  \\
        Fog                  & 44.49  & \textbf{45.00}  & \textbf{42.65}  & 41.24  & 44.04 & 51.15  \\
        Brightness           & \textbf{46.14}  & 45.24  & \textbf{43.91 } & 41.63  & 48.85 & 52.85  \\
        Contrast             & 14.98  & \textbf{17.15}  & 14.46  & \textbf{17.53}  & 21.09 & 27.79  \\
        Elastic Transform    & 29.34  & \textbf{39.29}  & 32.27  & \textbf{35.11}  & 36.41 & 38.80  \\
        Pixelate             & 31.51  & \textbf{38.02}  & \textbf{31.82}  & 30.10  & 46.26 & 49.92  \\
        JPEG Compression     & 15.86  & \textbf{27.55}  & \textbf{22.24}  & 21.87  & 37.35 & 40.35  \\ \midrule
        Average              & 20.60  & \textbf{28.50}  & 21.81  & \textbf{25.19}  & 28.41 & 31.36  \\ \bottomrule
    \end{tabular}
\end{table*}

\end{document}